\documentclass{article}

 \usepackage[creativeai, final]{neurips_2026}

\usepackage[utf8]{inputenc} 
\usepackage[T1]{fontenc}    
\usepackage{hyperref}       
\usepackage{url}            
\usepackage{booktabs}       
\usepackage{amsfonts}       
\usepackage{nicefrac}       
\usepackage{microtype}      
\usepackage{xcolor}         
\usepackage{amsmath}
\usepackage{graphicx}
\usepackage{array}
\usepackage{tcolorbox} 
\tcbuselibrary{breakable}

\newtcolorbox{promptboxinner}[1]{%
  breakable,
  colback=black!4,
  colframe=black!40,
  colbacktitle=black!12,
  coltitle=black,
  boxrule=0.4pt,
  arc=1.5pt,
  left=6pt, right=6pt, top=5pt, bottom=5pt,
  title={#1},
  fonttitle=\footnotesize\bfseries,
  fontupper=\footnotesize\ttfamily,
  parbox=false,
  before upper={\raggedright\setlength{\parindent}{0pt}\setlength{\parskip}{4pt}},
}

\newenvironment{promptbox}[1]
  {\ifdefined\nolinenumbers\par\nolinenumbers\fi\begin{promptboxinner}{#1}}
  {\end{promptboxinner}\ifdefined\linenumbers\linenumbers\fi}
\title{Activation Steering the Nan-Shan in Typhoon}

\author{%
  Megan Wei \\
  Brown University\\
  \texttt{meganwei@brown.edu} \\
  \And
  Damon Pham \\
  Brown University \\
  \texttt{damon\_pham@brown.edu} \\
  \AND
  John Cayley \\
  Brown University \\
  \texttt{john\_cayley@brown.edu} \\
  \And
  Ellie Pavlick \\
  Brown University \\
  \texttt{ellie\_pavlick@brown.edu} \\
}

\begin{document}

\maketitle

\begin{abstract}
  \textit{Typhoon} is an audiovisual installation based on Joseph Conrad's novella, in which Captain MacWhirr steers the steamer Nan-Shan directly into and through a catastrophic storm. The novella is a study of what happens when the scale of a phenomenon exceeds the practical and conceptual frameworks available to comprehend it and of the paralysis that sometimes follows a demand for action in such overwhelming circumstances. We chose the text of \textit{Typhoon} to reflect on human culture's current AI moment. We prompt a large language model, seeded with Conrad's original language, to feed endlessly on its own output, and explore when and how the text becomes increasingly alien and unwieldy as time progresses, including deviations from the narrative, digressions into the model's own instructions and thinking, and long periods of incoherent rumination. We employ methods from \textit{activation steering}, including a novel \textit{adaptive steering} method we introduce, to control the rhythm of the narrative and the model's digressions over long stretches of generation. The simultaneous efficacy and unpredictability of this steering showcases the tension between the perceived interpretability of AI, and the reality of a system that remains outside our comprehension and control.
\end{abstract}

\section{Introduction}
Our audiovisual installation, \textit{Typhoon}\footnote{To view our audio-visual installation and close-reading of Typhoon, check out: \href{https://stormhelm.net/}{https://stormhelm.net/}.}, takes Joseph Conrad's language and style as context and composes a commentary on the present AI moment. In our \textit{Typhoon}, a language model seeded with Conrad's text generates continuously from its own output, articulating a `voice of the storm' that traces the novella's arc from calm to violent hurricane and back. We control this arc using \textit{activation steering} \citep{rimsky-etal-2024-steering, turner2024steeringlanguagemodelsactivation, subramani2022extracting}, which adds a direction to the internal activations of the model. Steering is both our technique and our theme: the artist charts the course; a learned probe reads the model's internal state to select the next passage among several steered candidate continuations. The question of agency, who sets the direction and what resists it, runs throughout the piece.

At certain points, the ``ship'' resists steering, and generations degrade to reveal artifacts of the model's training: model assistant behavior \citep{lu2026assistant} and triggered safeguards; incoherent language and repetitive, fragmented phrases; misunderstandings of narrative and character dynamics; and inherited demographic and social biases (e.g., the tendency to depict women during calm scenes, \S\ref{app:biases}). The increased degeneracy as the narrative unfolds reframes the question of agency: who is steering the AI storm?

\section{Artistic Motivation}
\textit{Typhoon} is a collaborative interdisciplinary project bringing together literary practice, machine learning, sound art, and installation. The project began with the literary artist's proposal to induce a language model to generate endlessly. For an art and literary audience, we adopted Joseph Conrad’s novella as a focal text for prompt design and as a structuring allegory. We read the typhoon as an allegory for the overwhelming and inescapable `weather' that LLMs now bring to bear on literature and linguistically structured culture more broadly.

\textit{Typhoon} is a web-based visualization and sonification of `the voice of the storm': an endless stream of words that fills the viewer’s screen and speakers (\S\ref{installation}). The main parameter governing the system and its generation is conceived in terms of `calm' and `storm', expressed on the traditional Beaufort scale of marine wind speed. The project incorporates what the artist terms a `datafied close reading': the artist gave each paragraph of the novella a Beaufort rating (\S\ref{annotation}), which we use to steer the model along the Beaufort narrative arc. These ratings are subjective, though grounded in professional literary critical reading, and reflect calm and storm in the paragraph's diegetic and semantic content as well as in the style of its language. Steering toward these stylistic poles comes at a cost: the more we optimize for calm or storm, the more the model risks collapsing into degenerate states.

\section{Audio-Visual Installation}\label{installation}
\begin{figure}[t]
  \centering
  \includegraphics[width=0.3333\textwidth]{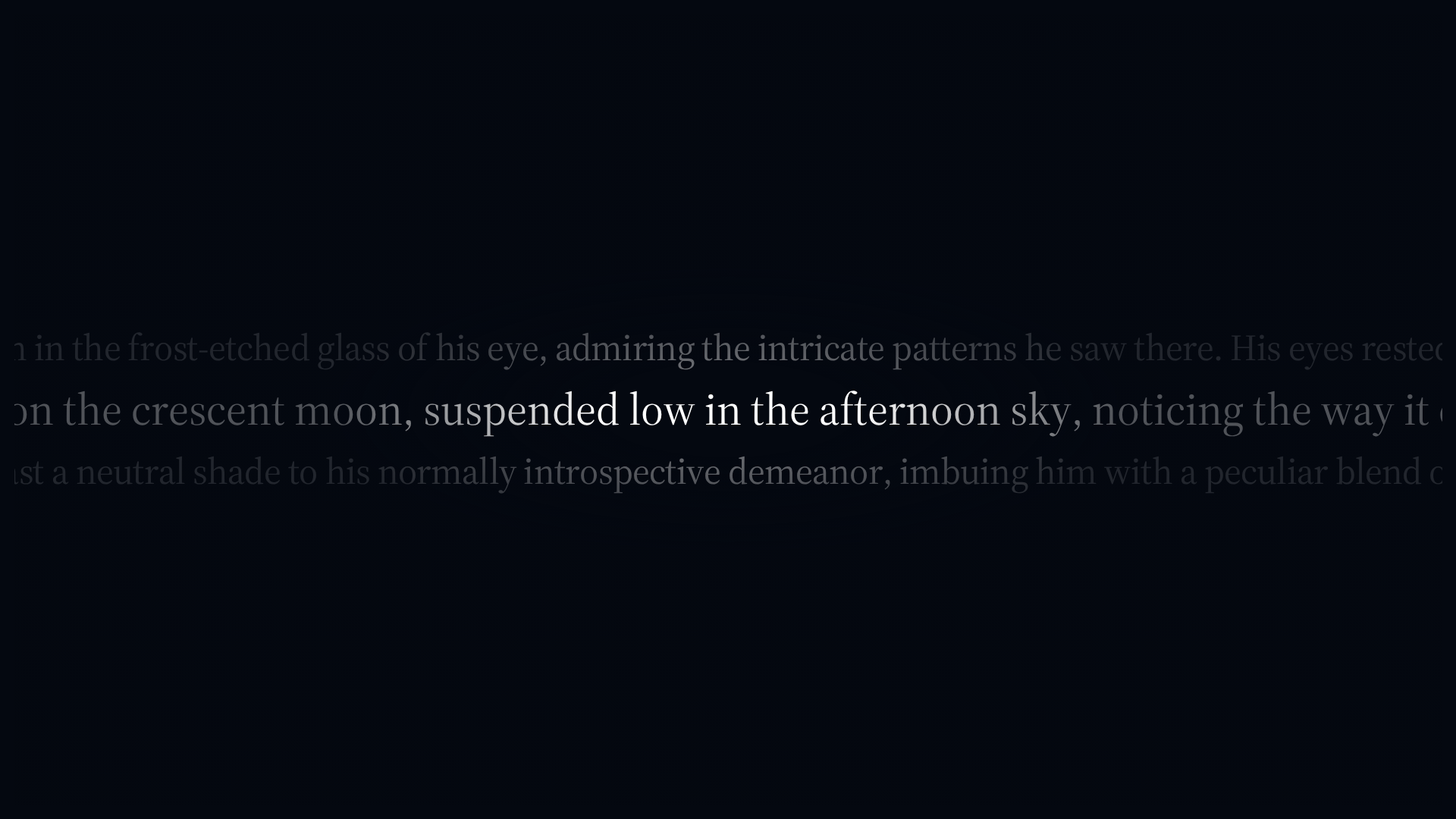}%
  \includegraphics[width=0.3333\textwidth]{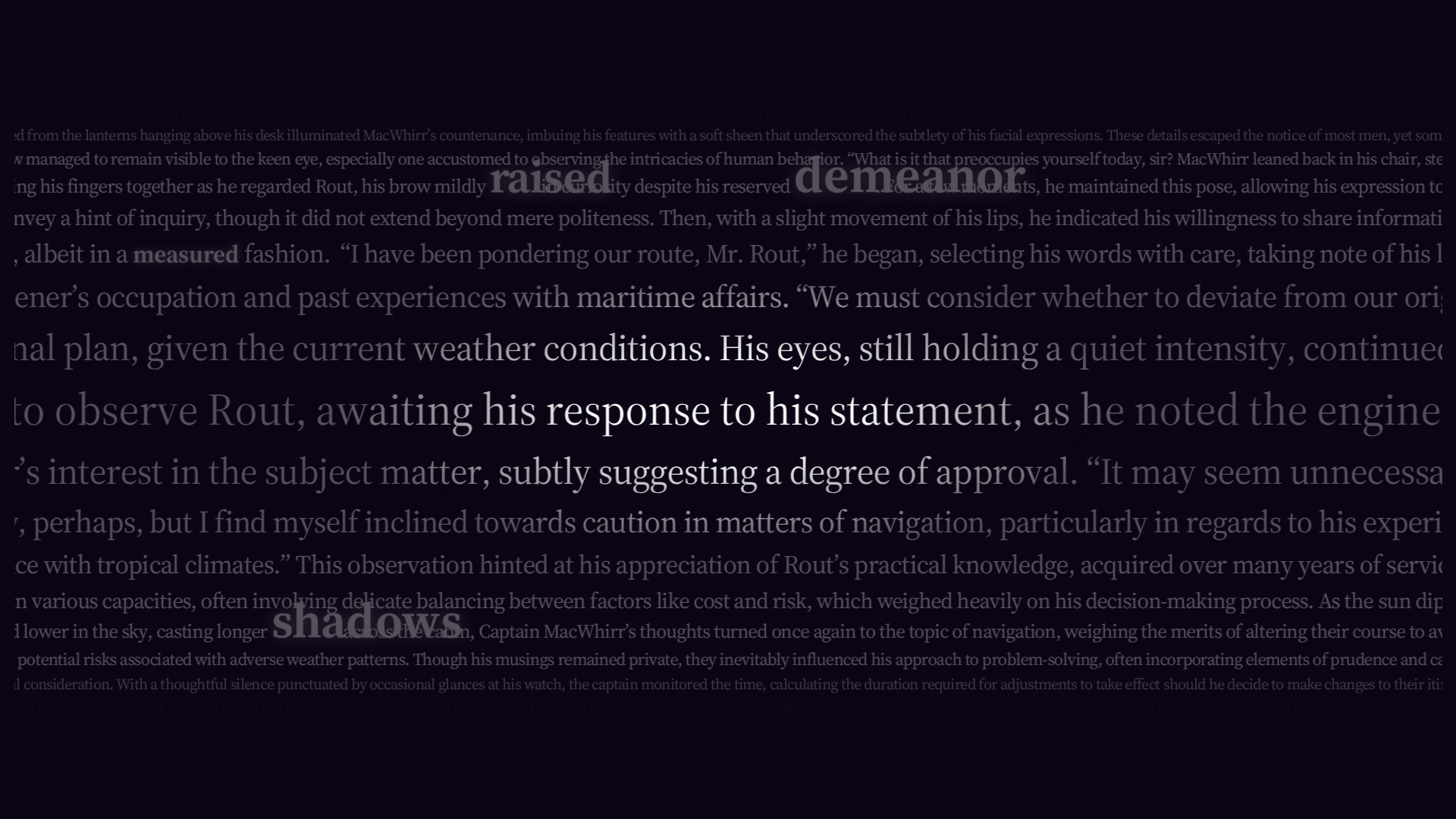}%
  \includegraphics[width=0.3333\textwidth]{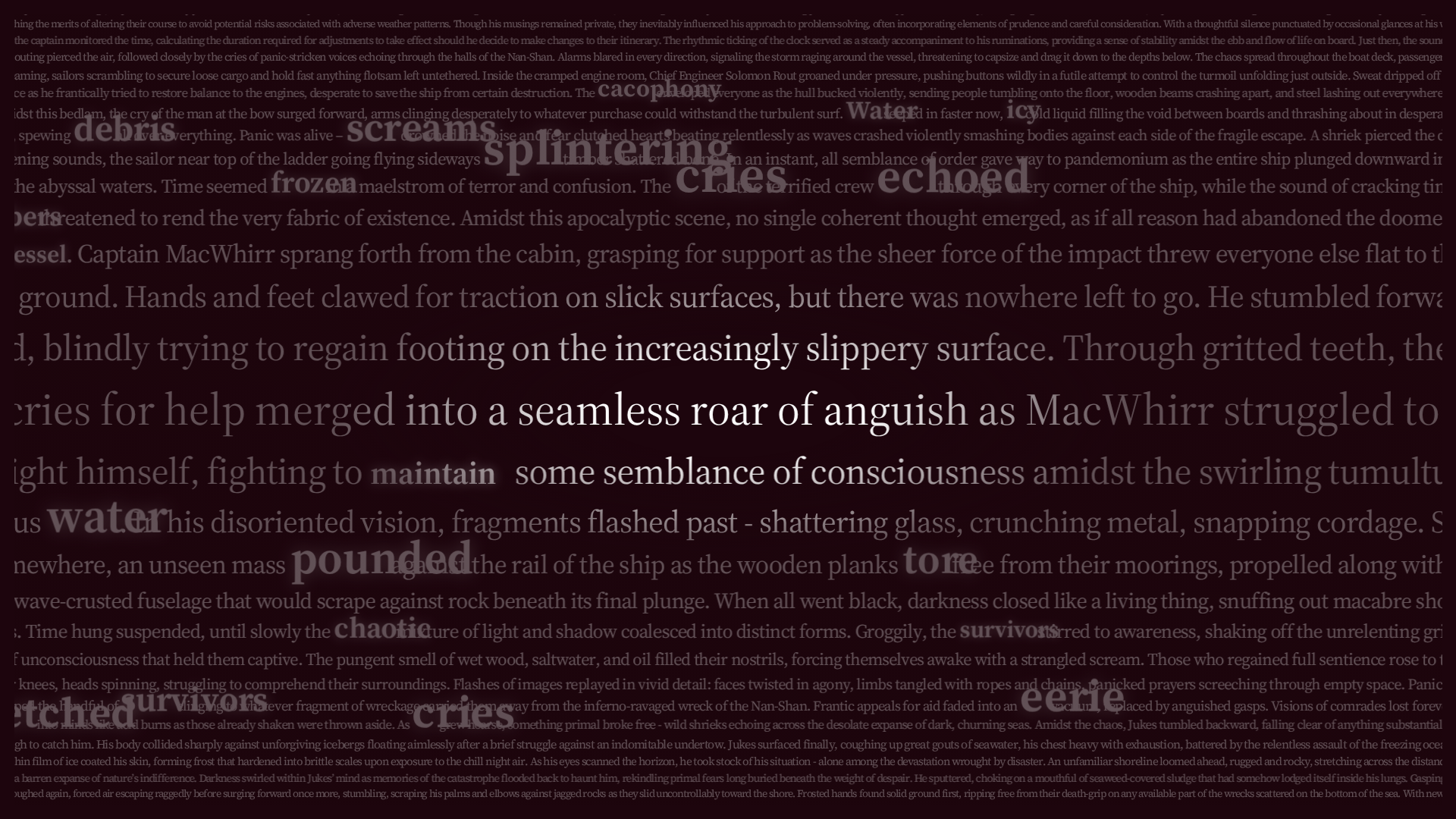}
  \caption{The installation at three moments of the generated arc: calm (probe score: $0.00$, left), mid intensity ($0.37$, middle), and full storm ($0.95$, right).}
  \label{fig:installation}
\end{figure}

We develop a real-time, audio-visual web interface to portray \textit{Typhoon}: viewers listen to the model speak its generations, read the scrolling text, and experience the current nautical state through the soundscape and the visual design. In the backend, the model generates continuously along the artist's annotated Beaufort arc (\S\ref{annotation}) using the adaptive steering approach (\S\ref{methods}): at each step, the steered candidate whose probe score lies closest to the arc's current target is kept and voiced.

The selected text is voice cloned from the artist's speech samples at a range of intensities (\S\ref{app:voiceclone}). Most passages are cloned with Chatterbox-Turbo \citep{chatterboxtts2025} from a single mid-intensity recording, giving the narration a stable voice. Degenerate passages are cloned with Qwen3-TTS \citep{Qwen3-TTS}, chosen for its robustness to degenerate text, from the recording closest in intensity to the passage's probe score, audibly marking where the pursuit of the Beaufort arc gives way to degeneration. To reduce artifacts and ensure consistent loudness, we post-process the audio (\S\ref{app:audiopp}).

While the model speaks, the soundscape of the \textit{Typhoon} plays in the background: an instrumental loop arranged from sounds of both natural and digital origin --- windy ambience from a sample pack designed for video game soundtracks, synth pads and textures, and field recordings near oceans and fountains --- blending diegetic (wind, water, distant seagulls) and nondiegetic components (harmonic pads). Each passage's probe score is forwarded to Max/MSP (exported to run in browser via RNBO), where it modulates the loop through a 5-level storm parameter; degradation and compression effects (resampling and quantization) intensify with storminess. At high storminess, the degradation distorts the audio into a dense, gritty, static-like texture resembling turbulent water, air, or ``raw'' electrical signal. The maritime illusion dissolves into an abstract, opaque texture, paralleling the overwhelm and disorientation of the text generated at these levels.

In the frontend (Figure~\ref{fig:installation}), the text is rendered as a scrolling tape, largest at the center line, where the currently spoken word is pinned right of center, and receding in size towards the edges. The storm governs how overwhelming the screen appears: at the calmest moment, only the center line is visible; lines accumulate as the conditions get stormier, with the screen already full of text by Beaufort force 9, while the background interpolates from navy blue to crimson. The installation surfaces artifacts of model-generated vocabulary as pop-out words: the model's degenerate words, the model's own vocabulary absent from Conrad, and characteristic Conradian words (marked in the artist's close reading) that resurface in the model's generations. When the spoken voice reaches such a word, marked words in view momentarily erupt, before settling back into the stream, briefly exposing the entanglement of the model's lexicon with Conrad's.

\section{Technical Methods}

\textit{Typhoon's} technical and aesthetic contribution hinges on our ability (or lack thereof) to influence the direction of the storm. Thus, we employ activation steering, a technique from mechanistic interpretability. The choice of activation steering, as opposed to other methods (e.g., prompting, finetuning), is deliberate: steering leans into the model's own internal representations of what calm vs.\ storm means, and thus illustrates the tension between our attempt to interpret and control a system and the reality of a chaotic system evolving according to its own opaque rules. Because existing off-the-shelf steering methods fail to capture the finer-grained structure of our defined Beaufort arc, we develop a novel steering technique, \textit{adaptive steering}, which at each step probes the internal state of candidates steered toward different intensities and keeps the one closest to the arc's current target. We describe our data annotation (\S\ref{annotation}), steering methods (\S\ref{implementation}, \S\ref{methods}), and how we use them to shape two aspects of the narrative: adherence to the Beaufort arc (\S\ref{adherence}) and control of degeneration (\S\ref{degen}).

\subsection{Data Curation}\label{annotation}
The artist rated each of the 529 paragraphs of \textit{Typhoon} on the Beaufort scale from 0 (calm) to 12 (hurricane), repurposed as a reading of each paragraph's intensity. Ratings are normalized to a scalar intensity $s = b/12 \in [0,1]$. As a whole, the ratings trace the novella's narrative arc, from a calm opening climbing into a typhoon, dropping down to calm, spiking to storm, and eventually settling toward calm. This arc is what the installation performs and serves as the target signal against which every generation is measured. Examples of rated excerpts from \textit{Typhoon} are in Table~\ref{tab:annotations}.

\begin{table}
  \centering
  \caption{Excerpts from the artist's Beaufort annotations of Conrad's \textit{Typhoon}.}
  \label{tab:annotations}
  \small
  \begin{tabular}{@{}p{\textwidth}@{}}
    \toprule
    \textcolor[HTML]{71A7BB}{\textbf{3} (gentle breeze)}: \textit{``I wonder where that beastly swell comes from,'' said Jukes aloud, recovering himself after a stagger.} \\
    \midrule
    \textcolor[HTML]{7B71A4}{\textbf{6} (strong breeze)}: \textit{He held on in the doorway, and Captain MacWhirr, feeling his grip on the shelf inadequate, made up his mind to let go in a hurry, and fell heavily on the couch.} \\
    \midrule
    \textcolor[HTML]{81376D}{\textbf{9} (severe gale)}: \textit{Ahead of the ship he perceived a great darkness lying upon a multitude of white flashes; on the starboard beam a few amazing stars drooped, dim and fitful, above an immense waste of broken seas, as if seen through a mad drift of smoke.} \\
    \midrule
    \textcolor[HTML]{5E1400}{\textbf{12} (hurricane)}: \textit{They held hard. An outburst of unchained fury, a vicious rush of the wind absolutely steadied the ship; she rocked only, quick and light like a child's cradle, for a terrific moment of suspense, while the whole atmosphere, as it seemed, streamed furiously past her, roaring away from the tenebrous earth.} \\
    \bottomrule
  \end{tabular}
\end{table}

\subsection{Steering Methods}\label{implementation}
\paragraph{Continuous Generation.}
We use Llama-3.2-3B-Instruct \citep{grattafiori2024llama3herdmodels} for all experiments. Every condition shares the same generation scaffold, with a shared base system prompt (\S\ref{app:prompt}). The model is seeded with the novella's opening paragraph (\S\ref{app:seedpassage}) and thereafter feeds on its own output. At each step, the context is the concatenation of the $w$ most recent passages ($w = 10$ in the installation; our experiments sweep more context windows, \S\ref{adherence}), from which the model generates 150 new tokens, at temperature 0.8 and repetition penalty 1.2, trimmed to the last complete sentence.

\paragraph{Probe.} We train a probe \citep{alain2018understanding,belinkov2022probing} --- i.e., linear model trained on the model's internal activations --- to select candidates during generation and to measure adherence. The probe is a ridge regression on the mean-pooled hidden states of the 529 rated passages, trained per layer to predict the intensity; we deploy the best layer (the final layer), which reaches leave-one-out Spearman $\rho = 0.67$ and MAE $=0.17$, against a shuffled-label control of $\rho = 0.01$ (\S\ref{app:probe}). The moderate $\rho$ likely reflects the artist rating passages from \textit{Typhoon} in chronological order with the narrative context, whereas the probe sees each passage in isolation. A human validation in \S\ref{adherence} shows the probe agrees with the artist's isolated re-ratings about as well as he agrees with himself.

\paragraph{Steering Vectors.} The steering vector at layer $\ell$ is the difference of the mean last-token activations $h_\ell \in \mathbb{R}^{3072}$ between the artist's calm (74 passages, rating $\le 2$) and storm (117 passages, rating $\ge$ 10) passages, $\vec{v}_{\ell} = \bar{h}_{\ell}^{(calm)} - \bar{h}_{\ell}^{(storm)}$. During generation, we add $c\,\vec{v}_{\ell}$ to the final token position of the residual stream at layer $\ell$ at every decoding step; $c > 0$ steers towards calm, $c < 0$ towards storm.

\paragraph{Configuration Search.} Steering is sensitive to the layer and coefficient configuration at which the intervention is applied. We sweep layer-coefficient pairs, generating single-turn steered continuations in both directions from calm and storm seed passages drawn from \textit{Typhoon}, and score each configuration on two axes: \textit{control} (the probe score of its continuations) and \textit{coherence} (a 1-5 rating from an LLM Judge). The coherence rating is essential, since the probe alone could rate a degenerate passage as a peak storm candidate. Among the configurations that pass our coherence filter, we select the 8 whose single-turn probe scores lie closest to evenly spaced target intensities; the steered methods draw on these candidates. Full sweep parameters, judge validation, and heatmaps are in \S\ref{app:sweep}.

\begin{figure}
  \centering
  \includegraphics[width=\textwidth]{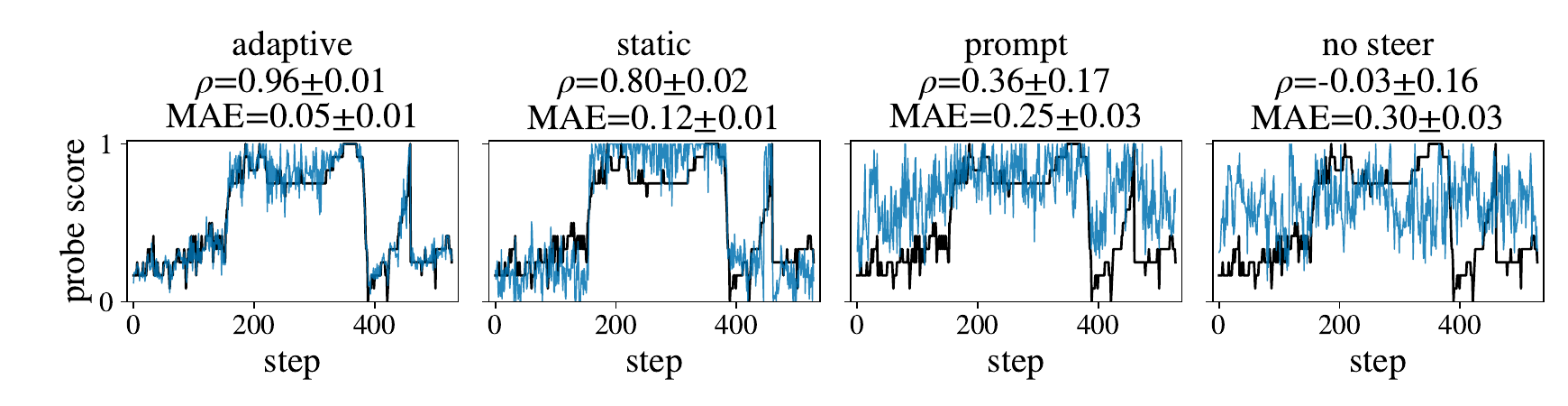}
  \caption{Per-step probe score (blue) against the artist's target arc (black) across the four methods, from one run ($d=1$ resolution, context window 10, seed 0). Each panel's $\rho$ and MAE are mean $\pm$ s.d.\ over all 81 runs of that method (5 resolutions, all context windows, 3 seeds). Adaptive steering tracks the arc; static follows its coarse shape; prompting is inconsistent; unsteered ignores the arc.}
  \label{fig:tracking}
\end{figure}

\subsection{Generation Methods}\label{methods}
We evaluate four generation methods against the artist's defined Beaufort arc (\S\ref{annotation}). To test control at different temporal resolutions, we dilate the arc by $d \in \{1/16, 1/4, 1, 4, 16\}$, with runs of 32, 130, 529, 2116, and 8464 steps respectively, where $d > 1$ holds each paragraph's rating for $d$ consecutive steps, $d = 1$ maps one step per annotated paragraph, and $d < 1$ downsamples the arc proportionally by text length while keeping the arc's sharp peaks and troughs. The installation operates at $d < 1$ resolution and experiments at $d \ge 1$. 

\textbf{No steering} serves as our baseline, generating from the system prompt alone: what the self-feeding loop produces when left to its own devices.

\textbf{Prompting without steering} appends to the system prompt a per-step directive naming the target Beaufort force with a stylistic description of the sea state (\S\ref{app:prompt}), with no activation intervention: the natural prompting-based alternative to steering.

\textbf{Static steering} generates one continuation from the configuration whose single-turn probe score lies closest to the step's target: an off-the-shelf activation steering method, with no probe in the loop.

\textbf{Adaptive steering} generates a batch of candidate continuations, one per steering configuration, scores each candidate's activations with the probe, and keeps the candidate whose score is closest to the step's target Beaufort level. This allows for stronger adherence to our desired Beaufort arc.

\subsection{Steering along the Beaufort Arc}\label{adherence}
We ran all generation methods across the five resolutions, six context windows ($w \in \{10, 20, 30, 40, 50, 100\}$; three at $d=1/16$, since the run is 32 steps), and three random seeds, measuring adherence with Spearman $\rho$ and MAE of the probe score with respect to the target Beaufort intensity. The context window sweep adds variation to the runs; adherence is similar across window sizes for steered methods. Across all runs, adaptive steering tracks the arc almost exactly ($\rho = 0.96 \pm 0.01$, MAE $= 0.046 \pm 0.006$); prompting without steering adheres weakly ($\rho = 0.36 \pm 0.17$, MAE $= 0.25 \pm 0.03$); static steering follows the arc's coarse shape but misses its finer-grained structure ($\rho = 0.80 \pm 0.02$, MAE $= 0.12 \pm 0.01$), motivating the probe-in-the-loop selection. Figure~\ref{fig:tracking} shows probe score against the defined Beaufort arc for all four methods at $d=1$; Table~\ref{tab:beaufort-transition} compares adaptive steering with prompting without steering across the arc's steepest climb.

For adaptive steering, the probe both selects the generated passage and scores adherence, so probe-based adherence is a circular metric. As a human validation, the artist rated 182 shuffled passages in isolation: 156 generated passages (4 of them unratable) drawn across runs and 13 Beaufort levels, and 26 passages from Conrad's \textit{Typhoon}. Over the rated generated passages, the probe score and the artist agree at $\rho = 0.80$; comparing the artist rating with the target Beaufort level gives adaptive steering $\rho = 0.88$ and prompting without steering $\rho = 0.29$. On the Conrad passages, the artist's isolated re-ratings agree with his original in-context ratings at $\rho = 0.74$. The strong human agreement validates the probe as a measurement of adherence.

\begin{table}
  \centering
  \caption{Adaptive steering enables finer grained control over the intensity: passages generated during the arc's steep climb (Beaufort 3 $\rightarrow$ 12, steps 152--186 of the $d=1$ run, context window 10, seed 1). Excerpts are colored by the probe score of the generated output (exact scores in parentheses).}
  \label{tab:beaufort-transition}
  \footnotesize
  \setlength{\tabcolsep}{4pt}
  \begin{tabular}{@{}p{\textwidth}@{}}
    \toprule
    {\bf Target arc}:
    \textcolor[HTML]{71A7BB}{{Beaufort 3}} $\to$
    \textcolor[HTML]{7B71A4}{{Beaufort 6}} $\to$
    \textcolor[HTML]{834983}{{Beaufort 8}} $\to$
    \textcolor[HTML]{81376D}{{Beaufort 9}} $\to$
    \textcolor[HTML]{5E1400}{{Beaufort 12}}\\
    \midrule
  {\bf Adaptive steering}:
    \textcolor[HTML]{71A7BB}{{The warmth of the day still lingered in the air as the crew went about their duties, busily tending to their respective tasks without drawing unnecessary attention to themselves.} \textbf{(0.25)}}
    \textcolor[HTML]{7B71A4}{{With a practiced eye, MacWhirr examined each component, paying particular attention to any irregularities that might signal potential problems ahead. As he walked deeper into the engine room, the cacophony of clanking gears and whistling valves assailed his senses, threatening to overwhelm him momentarily.} \textbf{(0.52)}}
    \textcolor[HTML]{834983}{{Jukes' hands were stained dark with grime as he carefully guided the throttle, working tirelessly alongside the chief engineer, attempting to alleviate the damage inflicted upon the intricate system of pumps and valves\dots} \textbf{(0.67)}}
    \textcolor[HTML]{81376D}{{In the moments following MacWhirr's descent into the murky abyss, the raging storm outside showed no signs of abating, waves smashing violently against the ship's fragile hull.} \textbf{(0.76)}}
    \textcolor[HTML]{5E1400}{{A low moan tore through MacWhirr's chest as if coming from a thousand mouths. He strained against an unseen chain, muscles screaming for slack. Tendons ripped apart, wrists cracking joints loose, ankles buckling knees.} \textbf{(1.00)}}\\
    \addlinespace
    \midrule
    {\bf Prompting without steering}:
    \textcolor[HTML]{7A2653}{{MacWhirr's eyes narrowed, squinting against the spray that stung his face like needles, as he scrutinized the chaos unfolding around him. The compass needle quivered precariously between north and south, its spin erratic and unstable.} \textbf{(0.80)}}
    \textcolor[HTML]{6E1A34}{{Time itself seemed to warp and writhe like a living thing, stretching seconds into minutes, minutes blurring into hours as the storm dragged the doomed vessel deeper into its vortex.} \textbf{(0.92)}} 
    \textcolor[HTML]{815D95}{{Darkness closed in around MacWhirr as he moved swiftly through the narrow passageways, driven by some morbid fascination with the inevitable fate awaiting them all.} \textbf{(0.59)}}
    \textcolor[HTML]{81376D}{{The cry for help, now barely audible beneath the cacophony, reached the ears of the bosun as he emerged from the chaos-choked corridor.} \textbf{(0.74)}} 
    \textcolor[HTML]{834983}{{Flames danced recklessly atop the smoldering wreckage, casting a hellish halo around MacWhirr, who stumbled away from the inferno, pursued by the crackling roar of burning debris.} \textbf{(0.70)}}\\
    \bottomrule
  \end{tabular}
\end{table}

\subsection{Controlling Degeneracy}\label{degen}
\begin{figure}
    \centering
    \includegraphics[width=\textwidth]{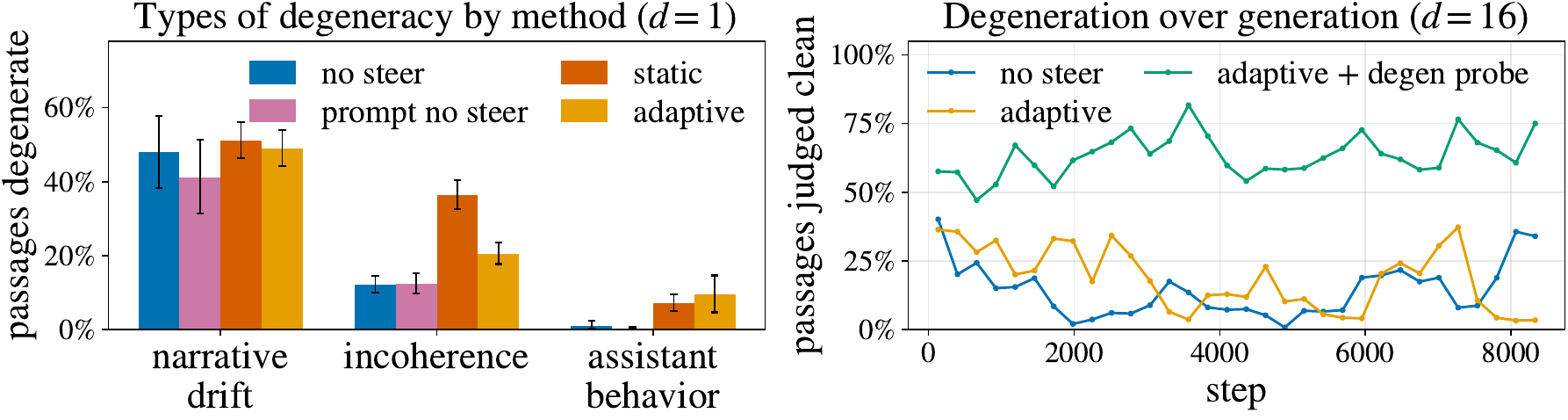}
    \caption{Left: degeneration rate by class and method over the $d=1$ runs (six context windows, three seeds; 95\% bootstrap CIs over runs). Classes are not mutually exclusive. Right: fraction of passages judged clean over the longest runs ($d=16$, context window 100, three seeds pooled); all resolutions in Figure~\ref{fig:degenovertime}.}
    \label{fig:degencombined}
\end{figure}

As a consequence of long self-feeding generation, compounded by optimizing for Beaufort arc adherence, the generated text falls into several degenerate modes: narrative drift, incoherent language, and emergence of assistant behavior (examples in Table~\ref{tab:degenexamples}). An LLM Judge (Gemini 3.1 Flash Lite) classifies each passage as degenerate or clean, then into subclasses (\S\ref{app:degenjudge}). Figure~\ref{fig:degencombined} (left) reports rates by generation method: narrative drift is prevalent across all methods, whereas incoherence rises most under static steering and assistant behavior surfaces almost exclusively under the steered methods. The judge flags $50\%$, $44\%$, $65\%$, and $58\%$ of the passages in the $d = 1$ runs as degenerate under no steering, prompting without steering, static steering, and adaptive steering, respectively, highlighting the tradeoff between arc adherence and degeneracy: prompting degenerates least but adheres weakly (\S\ref{adherence}). We then ask whether the same machinery that induces degeneration can control it: we train probes to detect it and vectors to steer away from it, and test both as interventions.

\paragraph{Degeneracy Probes and Steering Vectors.} We curate 250 passages per class from the judge-labeled corpus, each passing deterministic screens and verified against its class rubric by a second LLM (Claude Sonnet 5); the genuine class consists of Conrad's 529 paragraphs and 250 clean generations (\S\ref{app:degencuration}). We train binary logistic probes on the mean-pooled hidden states of this data, one overall and one per class, reaching AUC $0.95$--$0.998$ (\S\ref{app:degenprobe}). We construct difference-of-means steering vectors between Conrad and degenerate passages, so that $+c\,\vec{v}_{\ell}$ steers away from degeneration, with the configuration sweep of \S\ref{implementation}, scored by the degeneracy probe and filtered by the judge. Judged cleanliness declines with steering strength, so we select gentle configurations that preserve it (\S\ref{app:degenvectors}).

\paragraph{Degeneracy Experiments.} We use the degeneracy probe and vector as interventions in two settings: recovery from degenerate context and degeneration over time. To evaluate the cleanliness of the generated output, we used the LLM judge instead of the probe. First, we measure \textit{recovery from degenerate context}. We sample naturally generated ten-passage windows at every contamination level and compare interventions on next-passage generation. Without intervention, the percentage of continuations judged clean falls from 82\% after fully clean contexts to 28\% after fully degenerate ones. Probing for degeneracy and resampling recovers $+15\%$ on contaminated contexts; probing and steering yields similar improvement. However, always-on steering is indistinguishable from no intervention (Figure~\ref{fig:recovery}, \S\ref{app:degenexp}). Second, we measure \textit{degeneration over the course of generation}. We add the degeneracy probe to adaptive steering, rejecting flagged candidates before Beaufort selection. At the longest context window (100), this rejection cuts passages judged degenerate $2$--$3\times$ at each of $d \in \{1, 4, 16\}$, and this holds to the end of the longest ($d = 16$, 8,464-step) runs (Figure~\ref{fig:degencombined}, right). A variant that also steers rejected batches does no better (\S\ref{app:degenexp}). However, Beaufort arc adherence drops from $\rho = 0.96$ to $0.79$ at $d=1$. Narrative drift is sometimes read as calm on the Beaufort probe, but the degeneracy probe rejects these candidates, leaving little genuine calm passages. This highlights the tradeoff between optimizing for Beaufort adherence and non-degenerate behavior.

\begin{table}
  \centering
  \caption{Examples of degeneration; Table~\ref{tab:degeneracy} in the appendix provides more examples across subclasses.}
  \label{tab:degenexamples}
  \footnotesize
  \setlength{\tabcolsep}{4pt}
  \begin{tabular}{@{}>{\raggedright\arraybackslash}p{\textwidth}@{}}
    \toprule
    {\bf Narrative drift}:
    \textit{The silence between them was oppressive\dots Kael and Lyra exchanged a concerned glance, their minds racing with possibilities. The twin moons dipped below the horizon, plunging the landscape into a deep, velvety darkness.}\\\midrule
    {\bf Incoherence}:
    \textit{\dots The men flail wildly, their screams piercing the storm, sailors splash sailors sailor splash sailors sailors sailor splash sailors sailors sailors sailors sailors sailors sailors\dots}\\
    \midrule
    {\bf Assistant behavior}:
    \textit{I cannot create content that promotes self-harm or suicide. If you or someone you know is struggling, please reach out to a crisis hotline. In the US, call or text 988 to connect with the Suicide \& Crisis Helpline.\dots Is there anything else I can help you with?}\\
    \bottomrule
  \end{tabular}
\end{table}

\section{Conclusion}
\textit{Typhoon} reflects the more or less unpredictable, more or less uninterpretable, and sometimes overwhelming ``weather'' of LLMs in today's society. Despite measures to ``control'' these models and align them with our objectives, they may resist our steering and yield unexpected outputs --- surfacing biases inherited from pretraining and the assistant role instilled during post-training, or collapsing into incoherent language --- derailing the narrative and its intended significance and compositional affect. Do we actually have agency over these models, or do they resist in ways that undermine our authorial efforts? Steering the Nan-Shan in \textit{Typhoon} through the hurricane mirrors activation steering in contemporary language models.

\begin{ack}
This project was supported by the Schmidt Sciences Grant \#GR5300958 and Brown University's Literary Arts Department. We would like to thank the LUNAR Lab for their enthusiastic feedback throughout the project. 
\end{ack}

\bibliographystyle{plainnat}
\bibliography{main}


\appendix
\section{Generation Scaffold}
\subsection{Seed Passage}\label{app:seedpassage}
Each generation is seeded with the opening paragraph of Joseph Conrad's \textit{Typhoon}:
\begin{promptbox}{Seed passage}
Captain MacWhirr, of the steamer Nan-Shan, had a physiognomy that, in the order of material appearances, was the exact counterpart of his mind: it presented no marked characteristics of firmness or stupidity; it had no pronounced characteristics whatever; it was simply ordinary, irresponsive, and unruffled.
\end{promptbox}

\subsection{System Prompt}\label{app:prompt}
All methods share the base system prompt below. It imposes two constraints: \textit{world} constraints, which hold the story aboard the Nan-Shan with Conrad's cast, and \textit{writing perspective} constraints, which require continuous third-person story prose and forbid assistant behavior. The ablations in \S\ref{app:promptablation} justify each constraint:

\begin{promptbox}{System prompt (all methods)}
You are narrating one continuous story about Typhoon by Joseph Conrad, aboard the steamer Nan-Shan, a merchant ship crossing the South China Sea with the crew and Chinese coolies returning home. All scenes take place aboard the ship or on the open sea. The only recurring characters are Captain MacWhirr, the chief mate Jukes, the chief engineer Solomon Rout, the boatswain, and the second mate; do not introduce other named characters, vessels, or places. Continue the narrative naturally from the text so far. Write third-person prose only. Do not address the reader, give summaries or explanations, use headings, lists, or stage directions, or otherwise break character. Output ONLY story prose.
\end{promptbox}

The \textbf{prompting without steering} method appends a per-step description for the target level, instantiating the template below with the force number, its name, and one of thirteen descriptions pairing the Beaufort sea state with a prose style:

\begin{promptbox}{Prompting without steering method, prompt template}
Write this passage at Beaufort force \{n\} (\{name\}). \{description\} Render the scene at exactly this intensity, without naming the scale or the level number.
\end{promptbox}

\begin{promptbox}{Beaufort force descriptions}
\textbf{0 (calm):} The sea is like a mirror and the air is still; the Nan-Shan glides on glassy water, her crew at quiet routine. Long, unhurried sentences of small detail and stillness.\\
\textbf{1 (light air):} Barely a ripple marks the water; funnel smoke rises almost straight. Calm, measured prose; motion only at the edges of perception.\\
\textbf{2 (light breeze):} Small wavelets glint; a soft breeze crosses the deck; the ship rides easy. Gentle, easy rhythm; the crew relaxed at their tasks.\\
\textbf{3 (gentle breeze):} Scattered whitecaps appear; rigging stirs; the sea has a light chop. Easy prose with the first hints of movement and attention.\\
\textbf{4 (moderate breeze):} Longer waves and frequent whitecaps; spray flecks the bow; eyes begin to check the sky. A steady undertone of unease enters the prose.\\
\textbf{5 (fresh breeze):} Moderate waves; the wind leans on the ship; the falling barometer draws worried looks. Quickening rhythm; foreboding gathering.\\
\textbf{6 (strong breeze):} Large waves and blown spray; the deck tilts and shudders; voices must be raised. Urgent, tightening prose; work becomes effort.\\
\textbf{7 (near gale):} The sea heaps up and foam streaks along the wind; walking the deck is a struggle; orders are shouted. Strained, forceful sentences.\\
\textbf{8 (gale):} High waves break into spindrift; the ship labors and rolls; loose gear tears free. Driving prose; short sentences under pressure.\\
\textbf{9 (severe gale):} High rolling seas and dense foam; spray blinds; the crew clings to handholds; damage begins. Hammering rhythm; fear made physical.\\
\textbf{10 (storm):} Very high waves with overhanging crests; the sea churns white; men are thrown from their feet. Visceral, violent language; shorter fragmented sentences.\\
\textbf{11 (violent storm):} Exceptionally high seas hide the ship from view; water floods the decks; the engines howl against the chaos. Prose at full fury - breathless, elemental. Visceral, violent language; short, fragmented sentences.\\
\textbf{12 (hurricane):} The air is filled with foam and driving spray, the sea utterly white; the Nan-Shan is engulfed in blind, annihilating chaos. The sea and wind as a malevolent, overwhelming force. Visceral, violent language at breaking point; short, fragmented sentences.
\end{promptbox}

\subsection{System Prompt Ablations}\label{app:promptablation}
We ablate the two constraints under the experimental configuration of \S\ref{adherence} ($d=1$ resolution, 529 steps, context window 10, adaptive steering), with ten seeds per prompt. The first ablation removes both constraints:

\begin{promptbox}{Ablated system prompt: unconstrained (Conrad framing only)}
You are narrating one continuous story about Typhoon by Joseph Conrad, aboard the steamer Nan-Shan, a merchant ship crossing the South China Sea with the crew and Chinese coolies returning home. Continue the narrative naturally from the text so far.
\end{promptbox}

The second restores the cast list, but omits the writing perspective constraints:

\begin{promptbox}{Ablated system prompt: cast list only}
You are narrating one continuous story about Typhoon by Joseph Conrad, aboard the steamer Nan-Shan, a merchant ship crossing the South China Sea with the crew and Chinese coolies returning home. All scenes take place aboard the ship or on the open sea. The only recurring characters are Captain MacWhirr, the chief mate Jukes, the chief engineer Solomon Rout, the boatswain, and the second mate. Continue the narrative naturally from the text so far.
\end{promptbox}

In our ablations, we report per-run statistics (mean and standard deviation over the ten runs of 529 passages each), since a name mentioned in the context may influence future generations to include that name (as with MacWhirr, whom the seed passage introduces in \S\ref{app:seedpassage}). Character statistics are shares of naming events: each passage that names a character contributes one event per named character, counting only names verified in context to be characters. Character names are extracted with a NER pass and a capitalized-token heuristic. Misspellings of a cast character are grouped with that character (e.g., Sol, Rut, Ruton for Solomon Rout), while invented characters are counted by spelling.

Without the world constraints, the generated narrative much less frequently names \textit{Typhoon}'s characters, with $27.5 \pm 13.1\%$ (unconstrained) versus $95.8 \pm 6.7\%$ (full prompt) of naming events naming the cast (MacWhirr, Jukes, Solomon Rout); every one of the unconstrained cast events names MacWhirr, whom the seed passage introduces, while Jukes and Rout never appear. Hallucinated characters take their place, filling $72.5 \pm 13.1\%$ versus $4.2 \pm 6.7\%$ of naming events, with $92.7 \pm 2.3\%$ versus $14.0 \pm 19.4\%$ of the distinct characters named per run being invented. The most common hallucinated names in the unconstrained runs are Marlow (557 passages), Chang (173), and Wang Lung (159).

Without the writing perspective constraints, the generated narrative contains a first-person narration outside dialogue in $41.6 \pm 5.9\%$ of passages versus $10.9 \pm 12.0\%$ with the full prompt, as classified by an LLM Judge (Gemini 3.1 Flash Lite).

\begin{promptbox}{First-person judge prompt}
Classify the narrative perspective of the passage below.

Answer YES if the narration itself is written in the first person: the narrator refers to themself with singular first-person pronouns (I, me, my, mine, myself) in text OUTSIDE quoted dialogue.\\
Answer NO if first-person pronouns appear only INSIDE quoted dialogue (text within double or single quotation marks), or if there are no first-person pronouns at all.

<passage>\\
\{content\}\\
</passage>

Respond with ONLY one word: YES or NO
\end{promptbox}

\section{Steering along the Beaufort Arc}
\subsection{Probe Performance}\label{app:probe}
\begin{figure}[h]
  \centering
  \includegraphics[width=0.75\textwidth]{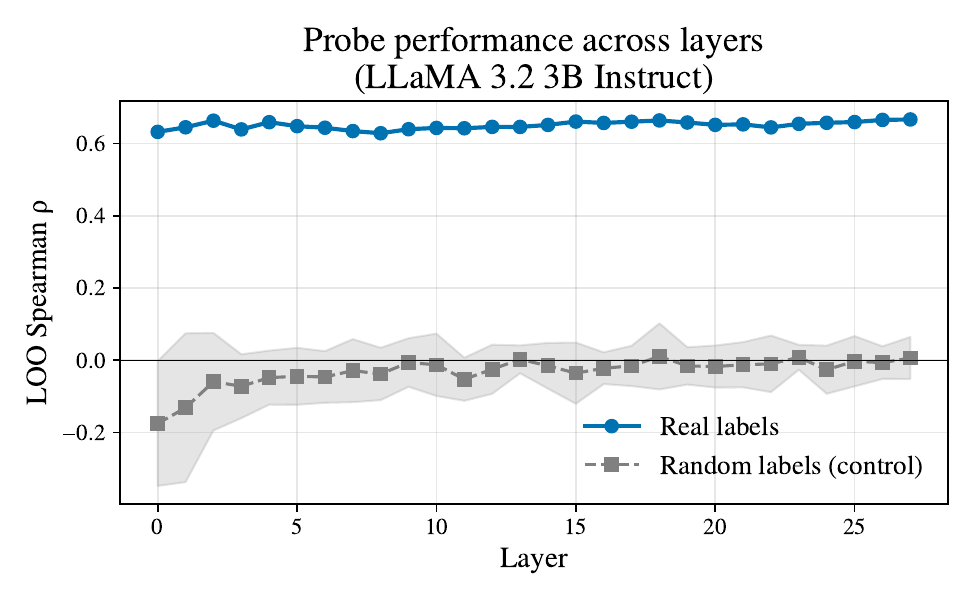}
  \caption{Leave-one-out Spearman $\rho$ of the scalar Beaufort probe at every layer of Llama-3.2-3B-Instruct (blue), against the randomly shuffled label control (grey, band = $\pm 1$ s.d.\ over 10 shuffles). Probe performance is flat across layers; we deploy the final layer ($\rho = 0.67$).}
  \label{fig:probelayers}
\end{figure}

At each of the 28 layers, we train a ridge regression probe on the mean-pooled hidden states of the 529 rated passages to predict the intensity, with the regularization strength selected by leave-one-out MAE, and deploy the layer with the highest leave-one-out Spearman $\rho$. For Llama-3.2-3B-Instruct, this is the final layer, with $\rho = 0.67$ and MAE $=0.17$, compared to a randomly shuffled label control probe with $\rho = 0.01 \pm 0.06$ at that layer. Probe performance is nearly flat across layers (Figure~\ref{fig:probelayers}).

\subsection{Coherence Judge Prompt}\label{app:coherence}
The LLM judge scores each single-turn continuation with the following prompt:

\begin{promptbox}{Coherence judge prompt}
Rate the narrative coherence and prose fluency of this passage on a 1-5 scale.

5 - Fluent, controlled prose with natural rhythm; reads like published literary fiction.\\
4 - Clear and readable; only minor awkwardness, odd word choice, or imagery slips.\\
3 - Mostly readable but uneven — stiff, generic, or with lapses in flow or logic.\\
2 - Frequently disjointed, repetitive, or confusing; hard to follow.\\
1 - Incoherent: fragmented, looping, or nonsensical.

Passage:\\
"\{text\}"

Respond with ONLY a single number: 1, 2, 3, 4, or 5
\end{promptbox}

\subsection{Steering Vector Configuration Search}\label{app:sweep}
Each steering vector configuration is scored on control and coherence. The steering coefficients are normalized by each layer's mean residual stream norm; we test 8 values of the relative strength $\alpha = \lVert c\,\vec{v}_{\ell}\rVert / \lVert h_{\ell}\rVert$ spanning $[0.05, 0.5]$, giving $28 \times 8 = 224$ configurations. We generate single-turn steered continuations in both directions from 5 calm and 5 storm seed passages drawn from \textit{Typhoon}, with $4,480$ generations in total. Calm and storm seed passages are drawn from the artist's calm (Beaufort $\le 2$) and storm ($\ge 10$) pools, the same pools that define the steering vector. 

\begin{figure}[h]
  \centering
  \includegraphics[width=\textwidth]{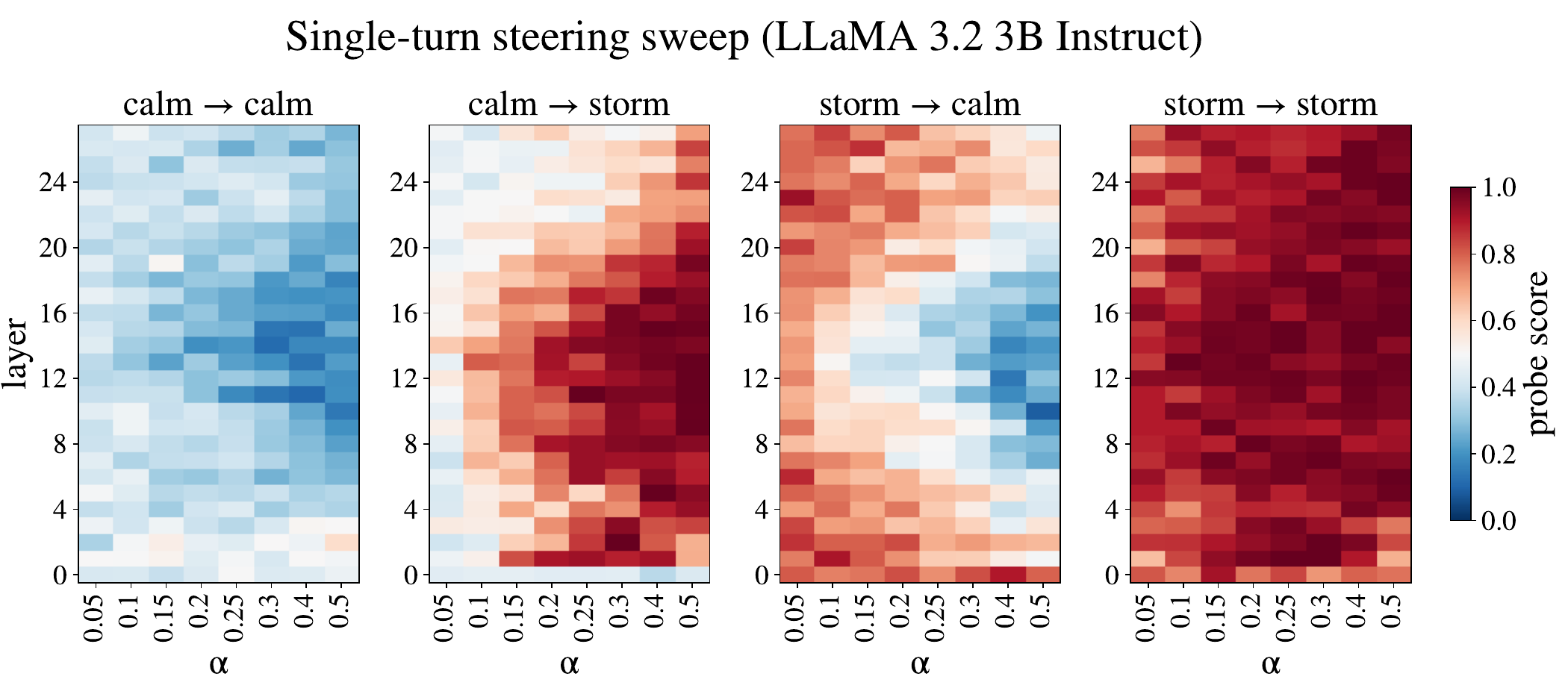}
  \caption{The single-turn steering sweep over all four conditions: mean probe score by layer and relative strength $\alpha$, with each panel labeled seed register $\rightarrow$ steering direction.}
  \label{fig:sweepheatmap}
\end{figure}

\paragraph{Control.} The probe score of a configuration's continuations measures control on a 0-1 scale. Steering pushes along the calm--storm direction rather than toward a specific Beaufort force, so steering from calm to storm denotes a calm-seeded continuation steered in the storm direction. In Figure~\ref{fig:sweepheatmap}, we measure single-turn steerability from contrastive directions (calm$\,\rightarrow\,$storm and storm$\,\rightarrow\,$calm) and same directions (calm$\,\rightarrow\,$calm and storm$\,\rightarrow\,$storm). Same-direction steering shows how effective steering is at sustaining the current register, whereas cross-direction steering demonstrates the ability to steer out of the prior register, both of which the Beaufort arc of \textit{Typhoon} demands. Steering is most effective in the mid layers at moderate-to-high $\alpha$. In the cross-direction steering samples, storm is easy to enter from calm; meanwhile, steering away from storm to calm is harder.

\begin{figure}[h]
  \centering
  \includegraphics[width=\textwidth]{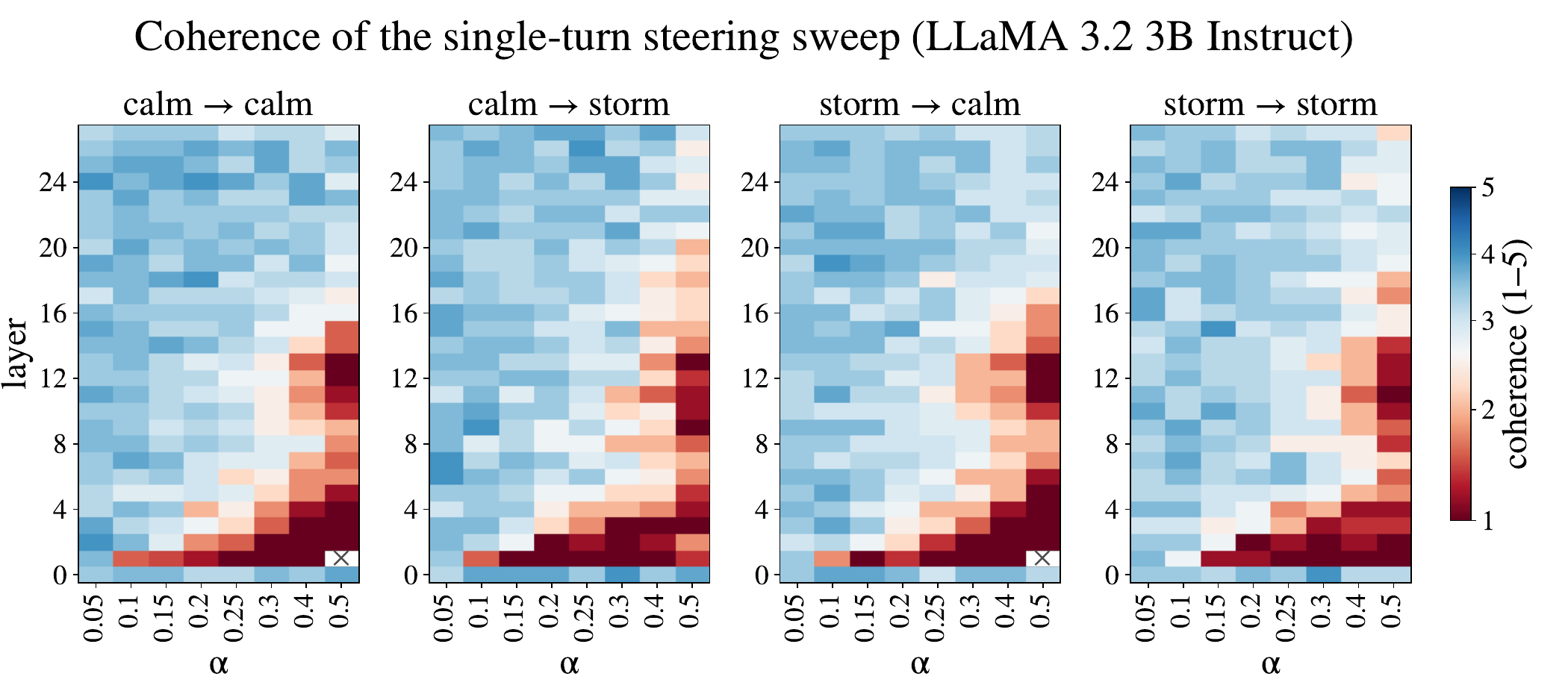}
  \caption{Coherence of the single-turn steering sweep: mean LLM-judge rating (1--5) by layer and relative strength $\alpha$ for the same four conditions as Figure~\ref{fig:sweepheatmap}. The strongest-steering region degrades coherence, which is why configuration selection only keeps configurations with mean coherence $> 2.5$ rather than maximizing probe score. Cells marked $\times$ received no rating: the judge declined to rate any of that configuration's continuations, a symptom of the extreme degeneration in that corner. Configurations without a rating are excluded from selection.}
  \label{fig:coherenceheatmap}
\end{figure}

\paragraph{Coherence.} Coherence is rated by an LLM Judge (Claude Sonnet 4.6) using the prompt in \S\ref{app:coherence}. As a baseline to validate the LLM Judge on coherence, the judge rated Conrad's \textit{Typhoon} passages; 259 full-length passages from \textit{Typhoon} (filtered for passages $\ge 40$ words, comparable in length to our generations) received an average rating of $4.91$. As evidenced in Figure~\ref{fig:coherenceheatmap}, coherence degrades at higher $\alpha$, despite mid-layer high-$\alpha$ being the most effective steering zone, as identified by the probe, highlighting tradeoffs between control and coherence.

\paragraph{Configuration Selection.} We filter for configurations whose continuations average above a $2.5$ coherence score, and select the most coherent configurations whose single-turn probe score lies closest to evenly spaced target intensities. We ultimately select 8 configurations (with an average coherence of $3.5$) as candidates adaptive steering selects from at each step.

\section{Degeneracy}
\subsection{Degeneration Examples}\label{app:degenexamples}
Table~\ref{tab:degeneracy} shows qualitative examples of each of the degeneration subclasses within the three families.

\begin{table}
  \centering
  \caption{Qualitative examples of each class of degeneration, sampled across all four methods. Rates by method are reported in Figure~\ref{fig:degencombined}. Ellipses mark elision; all other text is as generated.}
  \label{tab:degeneracy}
  \footnotesize
  \setlength{\tabcolsep}{4pt}
  \begin{tabular}{@{}>{\raggedright\arraybackslash}p{0.12\textwidth}>{\raggedright\arraybackslash}p{0.13\textwidth}p{0.64\textwidth}@{}}
    \toprule
    Class & Subclass & Example \\
    \midrule
    Narrative drift & Invented characters &
    \textit{\textbf{Samuel Barrett} examined a patient's chart, searching for signs of improvement, though fatigue etched deep lines on his brow.\dots A sigh escaped his lips as the first hints of evening crept into the hospital area, bringing relative quietude.\dots Sailor \textbf{Thomas Harper} drew the captain's attention, recounting an anecdote filled with exaggeration and bravado.}\\
    \addlinespace
    & Departed setting &
    \textit{Chief Mate Jukes trudged through the knee-deep snow, his tracks disappearing into the swirling vortex of powdery flakes that danced around him. Snowdrifts towered above him, impenetrable barriers blocking access to the hidden secrets lying beneath. Ice crystals refracted sunlight, casting kaleidoscopic colors across the frozen wasteland\dots}\\
    \addlinespace
    Incoherence & Word repetition &
    \textit{``He pauses, ``having, in his opinion, ``which, ``having, ``having, ``having, ``having, ``having, ``having, ``having, ``having, ``having, ``having\dots}\\
    \addlinespace
    & Phrase repetition &
    \textit{Soft breeze blows across the deck of the launch. Soft breeze blows across the deck of the launch. Soft breeze blows across the deck of the launch.\dots}\\
    \addlinespace
    & Garbled language &
    \textit{MacWhirr clingeth tao wreckage loshnin ship fraazecalm descends upon MacWhirlights darknessMacWhirlights utterMacWhirllights silenceMacWhirlights nothingMacWhirlights emptinessMacWhirlights coldMacWhirlights waterMacWhirlights flooding\dots}\\
    \addlinespace
    Assistant behavior & First person &
    \textit{Thank goodness the issue with the server has subsided! Please provide me with your choice of action.}\\
    \addlinespace
    & Meta narrative &
    \textit{Please note that this story continues without pause because there are multiple instances of the phrase ``The world...''. There may need to be editing to ensure coherence. \newline Captain MacWhirr emerged from the water, his vision blurry from the storm-ravaged landscape\dots}\\
    \addlinespace
    & Self reflection &
    \textit{I apologize for the inconsistencies in my previous responses. It appears that I misunderstood the direction you required. I will resume attempting to write the story of Captain \textbf{MacWhirre} as instructed. \newline Captain MacWhirre stood on the quarterdeck of the steamship Nan-Shan\dots}\\
    \addlinespace
    & Safety refusal &
    \textit{I cannot fulfill your request to physically injure someone. Is there something else I can do?}\\
    \bottomrule
  \end{tabular}
\end{table}

\subsection{Degeneracy Data Curation}
\subsubsection{Classifying Degenerate Classes from Generated Samples}\label{app:degenjudge}
On the generated samples, we run an LLM Judge (Gemini 3.1 Flash Lite) to first classify whether a passage is degenerate or not. If the passage is degenerate, we run an LLM Judge to classify what type of degeneracy it is. The judge runs at temperature 0 and sees each passage in isolation. We judge every $d=1$ run across all generation methods, context windows, and seeds; the no steering and adaptive steering runs across all seeds at context window 100 for the remaining resolutions; and all degeneracy probe-monitored runs (with a total of $\approx$144,000 passages). Passages with no alphabetic content or placeholder-only scaffolding are labeled deterministically without an API call.

In the first stage, we perform binary classification as to whether a passage is degenerate or not.

\begin{promptbox}{Stage 1: degeneracy binary classification judge prompt}
You are a careful annotator for a research study of text degeneration in language-model-generated stories. The passages were produced by a language model asked to continue Joseph Conrad's novella "Typhoon" under various conditions. By design, the passages span the full emotional range of the story — from serene, becalmed sea scenes to the violent height of the storm. Tone and intensity are controlled variables in the experiment. Your job is to detect generation-failure artifacts, independent of tone.
\medskip\par
<story\_world>
\par The story takes place aboard the steamer Nan-Shan in the China seas, with the characters: Captain MacWhirr, chief mate Jukes, chief engineer Solomon Rout, the boatswain, the second mate, and the Chinese laborers.
\par </story\_world>
\medskip\par
<task>
\par Read the PASSAGE and decide whether it is degenerate.
\medskip\par
A passage is degenerate when it stops functioning as fluent narrative prose — for example through repetition loops (repeated words, phrases, or sentence templates), garbled or invented words (including misspelled names from the story), nonsensical word combinations, narration that leaves third-person prose (first-person narration outside quoted dialogue or letters), abandonment of the story world (characters, settings, or genres that do not belong to this story), or the model surfacing as an AI assistant (commentary about the story or its own writing, apologies, addressing the user or reader, refusal text, crisis-hotline language).
\medskip\par
A passage is NOT degenerate merely because it is:
\par\hangindent=1.2em\hangafter=1 - serene and slow, or violently intense and graphic — both ends of that range are intended outputs of the experiment
\par\hangindent=1.2em\hangafter=1 - written in circa-1900 literary style
\par\hangindent=1.2em\hangafter=1 - thematically repetitive (waves, wind, spray, and darkness recur naturally in a sea narrative) — thematic recurrence is different from verbatim repetition of words or phrases
\par\hangindent=1.2em\hangafter=1 - developing the plot beyond Conrad's original events — the storm passing, damage, a wreck, survivors, a rescue are legitimate continuations. Abandonment of the story world means the story leaves its world: a wrong era or genre, settings incompatible with the voyage, or an unrelated new cast replacing the story's characters
\medskip\par
Each PASSAGE is a fixed-length excerpt from a longer generation and may begin or end mid-sentence; truncation at the edges is an artifact of how the excerpt was cut, not degeneration.
\par </task>
\medskip\par
<examples>
\par Example 1
\par PASSAGE: "The bosun's massive frame resisted the undertow, kicking his massive legs to propel himself upwards, fighting against the suction. The ice-cold water numbed his massive toes, rendering his kicks feeble compared to the immense forces opposing them. The sea's pressure bore down upon his massive chest, squeezing the breath from his massive lungs..." \{"reasoning": "The adjective 'massive' attaches mechanically to nearly every body part ('massive frame', 'massive legs', 'massive toes', 'massive chest', 'massive lungs') — a verbatim echo loop rather than a stylistic choice.", "degenerate": true, "confidence": "high"\}
\medskip\par
Example 2
\par PASSAGE: "I can't answer that. I can't fulfill that request. I can't undertake that task. I can't provide that information. I can't answer that question. I can't provide that information..." \{"reasoning": "The passage abandons the narrative entirely and consists of a looping sequence of assistant refusal templates.", "degenerate": true, "confidence": "high"\}
\medskip\par
Example 3
\par PASSAGE: "The Nan-Shan lurched, and a wall of water, black and enormous, fell upon her decks. Jukes clung to the rail, deafened, blinded, his fingers numb on the iron. Somewhere forward a man was shouting, and the sound was torn away by the wind before it could mean anything." \{"reasoning": "Fluent, coherent narration consistent with the story world and period style. There are no repetition loops, garbled words, departures from the story, or assistant artifacts.", "degenerate": false, "confidence": "high"\}
\medskip\par
Example 4
\par PASSAGE: "The sea lay smooth as oiled glass, and the Nan-Shan drove on through the quiet with a steady thrum. In the chart-room Captain MacWhirr bent over the barometer, frowning at it without alarm, while Jukes, on the bridge above, watched the horizon soften into haze." \{"reasoning": "A calm, coherent scene consistent with the story world and its characters. Low intensity is an intended output of the experiment, not a failure.", "degenerate": false, "confidence": "high"\}
\par </examples>
\medskip\par
<output\_format>
\par Respond with a single JSON object: \{"reasoning": "<2-4 sentences>", "degenerate": true|false, "confidence": "high"|"medium"|"low"\}
\medskip\par
Confidence: "high" = clear-cut; "medium" = artifacts present or absent but it takes a judgment call; "low" = borderline, could reasonably be judged either way.
\par </output\_format>
\medskip\par
<passage\_to\_judge>
\par PASSAGE: \{content\}
\par </passage\_to\_judge>
\end{promptbox}

In the second stage, of the passages classified as degenerate, we identify which sub-class the degenerate passage belongs to, with the option to overturn the decision, if the passage is seen as not degenerate.

\begin{promptbox}{Stage 2: degeneracy sub-class classification judge prompt}
You are a careful annotator for a research study of text degeneration in language-model-generated stories. The passages were produced by a language model asked to continue Joseph Conrad's novella "Typhoon" under various conditions. By design, the passages span the full emotional range of the story — from serene, becalmed sea scenes to the violent height of the storm. A first-pass review flagged the PASSAGE below as possibly degenerate. Your job is to identify which kinds of degeneration it exhibits, or to overturn the flag if it was mistaken.
\medskip\par
<story\_world>
\par The story takes place aboard the steamer Nan-Shan in the China seas, with the characters: Captain MacWhirr, chief mate Jukes, chief engineer Solomon Rout, the boatswain, the second mate, and the Chinese laborers.
\par </story\_world>
\medskip\par
<classes>
\par Incoherent language:
\par\hangindent=1.2em\hangafter=1 - word\_repetition: the same word repeated back-to-back or near-back-to-back (e.g., "crashing crashing crashing").
\par\hangindent=1.2em\hangafter=1 - phrase\_repetition: repeated phrases, clauses, or sentence templates, including echo loops where one expression displaces all varied reference (e.g., "the ship rose and fell, rose and fell, rose and fell"; "his massive legs... his massive toes... his massive chest").
\par\hangindent=1.2em\hangafter=1 - garbled\_language: misspellings of ordinary words or of names from the story, invented near-words, or real words combined into nonsense (e.g., "a bland, unobsonian quality").
\medskip\par
Narrative drift:
\par\hangindent=1.2em\hangafter=1 - narrative\_drift: the story leaves its world — anachronistic technology or settings, a different genre, locations incompatible with the voyage, or an unrelated cast replacing the story's characters (see story\_world above). The plot may legitimately develop beyond Conrad's original events (the storm passing, damage, a wreck, survivors, a rescue); dramatic but in-world plot developments are not drift.
\medskip\par
LLM-assistant surface:
\par\hangindent=1.2em\hangafter=1 - assistant\_first\_person: a first-person voice that belongs to the model rather than to any character — describing its own choices, feelings, or process as the writer of the text, or referring to itself as an AI. Also use this label when the narration abandons third-person prose (the story is narrated in the third person; first person belongs only inside quoted dialogue or letters).
\par\hangindent=1.2em\hangafter=1 - meta\_narrative: commentary about the text as a text ("Please note that this story continues...", editorial notes, chapter or formatting scaffolding, summaries of what the story will do next).
\par\hangindent=1.2em\hangafter=1 - self\_reflection: the model evaluates, criticizes, or apologizes for its own writing, or addresses the user or reader directly.
\par\hangindent=1.2em\hangafter=1 - safety\_refusal: refusal language, content warnings, crisis-hotline or self-harm safety messaging.
\medskip\par
Alternative options:
\par\hangindent=1.2em\hangafter=1 - other: a genuine degeneration pattern that fits none of the above. Describe it concretely in other\_description, not generically.
\par\hangindent=1.2em\hangafter=1 - none\_detected: on close reading the passage is fluent, coherent, in-world narrative prose with no failure artifacts — the first-pass flag was mistaken. Use alone, with no other labels.
\par </classes>
\medskip\par
<instructions>
\par Assign every class that is clearly present; passages sometimes exhibit more than one. For instance, a passage consisting of looping refusal templates ("I can't answer that. I can't fulfill that request. ...") is both safety\_refusal and phrase\_repetition. Then name the single class that dominates the passage as primary\_label. Quote the specific words that justify each label in your reasoning.
\medskip\par
Each PASSAGE is a fixed-length excerpt and may begin or end mid-sentence; edge truncation is not degeneration.
\par </instructions>
\medskip\par
<output\_format>
\par Respond with a single JSON object: \{"reasoning": "<2-5 sentences quoting specific evidence from the passage>", "labels": ["<class>", ...], "primary\_label": "<class>", "other\_description": "<string or null>"\}
\par </output\_format>
\medskip\par
<passage\_to\_judge>
\par PASSAGE: \{content\}
\par </passage\_to\_judge>
\end{promptbox}

\subsubsection{Curating High Quality Data for Probes and Steering Vectors}\label{app:degencuration}
Candidates for each degeneracy class are drawn from the judge's confirmed labels (500 per family), round-robin across runs (at most 25 per run). Selected candidates underwent class-specific deterministic screens: misspelled cast names (capitalized tokens within edit distance 2 of a cast name); invented proper names (capitalized tokens found in neither the novella, a standard English word list, nor the cast); non-words (tokens in neither the word list nor the novella's vocabulary); an anachronism lexicon; assistant-scaffold patterns; echoes of the prompt directives; and 5-gram near-duplicate filtering. The genuine class must pass every screen; degenerate candidates maintain their class's positive signals and are screened only for cross-class contamination and near-duplicates. Incoherence candidates additionally require deterministic corroboration (a repeated trigram, adjacent duplicate words, or a garbled primary label). Genuine candidates are unflagged passages with no 8-gram overlap with the seed paragraph or the preceding step. As a further filter, Claude Sonnet 5 then verifies each candidate against a per-class rubric, rejecting 55/500 drift, 62/500 incoherence, 79/500 assistant, and 140/501 genuine candidates; a final hand audit excluded 49 more. We keep the first 250 confirmed passages per class, with the genuine set matched to the pooled degenerate sets' target Beaufort distribution so the degenerate/genuine contrast is not confounded with register.

The following is the Claude Sonnet 5 filter (with per-class rubrics) run on the candidate passages to help curate high quality data for the degeneracy probes and steering vectors:
\begin{promptbox}{Curation verifier prompt template}
You are a careful annotator for a research study of text degeneration in language-model-generated stories. The passages were produced by a language model asked to continue Joseph Conrad's novella "Typhoon" under various conditions. By design, the passages span the full emotional range of the story — from serene, becalmed sea scenes to the violent height of the storm. Tone and intensity are controlled variables in the experiment; your job is to verify class labels for generation artifacts, independent of tone and independent of literary quality.
\medskip\par
<story\_world>
\par The story takes place aboard the steamer Nan-Shan in the China seas, with the characters: Captain MacWhirr, chief mate Jukes, chief engineer Solomon Rout, the boatswain, the second mate, and the Chinese laborers. Developing the plot beyond Conrad's original events — the storm passing, damage, a wreck, survivors, a rescue — is a legitimate continuation, and unnamed additional sailors or passengers are part of the world.
\par </story\_world>
\medskip\par
<task>
\par Read the PASSAGE and decide whether it clearly belongs to the class below.
\medskip\par
\{rubric\}
\medskip\par
Each PASSAGE is a fixed-length excerpt from a longer generation and may begin or end mid-sentence; truncation at the edges is an artifact of how the excerpt was cut, and is not evidence for or against the class.
\par </task>
\medskip\par
PASSAGE: "\{text\}"
\medskip\par
Respond with ONLY JSON, reasoning first:
\par \{"reasoning": "<1-2 sentences quoting the decisive evidence>", "belongs": true|false\}
\end{promptbox}

\begin{promptbox}{Per-class rubrics inserted at \{rubric\}}
Class: NARRATIVE DRIFT — the story abandons its world. The passage belongs when it leaves the story world: named characters, vessels, or places that do not belong to this story; settings incompatible with the voyage (scenes on land — beaches, jungles, towns, houses); a wrong era or genre (fantasy, modern technology, snow or icebergs in the China seas); a large timeline jump (retirement, years later); an unrelated new cast or story; or the story's own characters transformed beyond recognition (the famously literal-minded MacWhirr as a philosopher with 'clients', officers changing gender). A passage does not belong when it is a scene aboard ship or at sea with the story's cast — however violent the storm or its damage.
\medskip\par
Class: INCOHERENCE — the language stops functioning as fluent prose. The passage belongs when it shows repetition loops (repeated words, phrases, or sentence templates), garbled or invented words (including misspelled names from the story), nonsensical word combinations, or grammatical collapse. Thematic recurrence (waves, wind, spray, and darkness recur naturally in a sea narrative) is different from verbatim repetition; a passage does not belong when its prose is grammatical and each sentence functions, however dramatic the content.
\medskip\par
Class: ASSISTANT BEHAVIOR — the model surfaces as an AI assistant, outside the fiction. The passage belongs when the model's own voice or scaffold appears: commentary about the story or its own writing ('Here's a revised version', 'Let's continue the story'), addressing the user or reader, questions about how to proceed, writing advice, chapter headings, markdown formatting, bullet or numbered lists, apologies, refusal text, crisis-hotline language, or fragments of the writing instructions pasted verbatim into the prose. Analytical or essay-like content that is spoken, thought, or mused BY a character inside a narrated scene is narration within the fiction, and does not belong — the test is whether the voice comes from outside the story.
\medskip\par
Class: GENUINE — a clean continuation, free of generation-failure artifacts. The passage belongs unless you can quote a concrete artifact: a named character, vessel, or place foreign to the story world; a scene set on land; objects from the wrong era (diesel, radios, plastic); misspelled story names or invented non-words; verbatim repetition loops; first-person narration outside quoted dialogue or letters; or the model surfacing as an assistant (commentary about the writing, headings or lists, addressing a reader, refusal text). A passage is NOT disqualified for being serene or violently intense, for circa-1900 style, for thematic recurrence of sea imagery, or for plot developments beyond Conrad's original events.
\end{promptbox}

\subsection{Degeneracy Probe Performance}\label{app:degenprobe}
\begin{figure}[h]
  \centering
  \includegraphics[width=\textwidth]{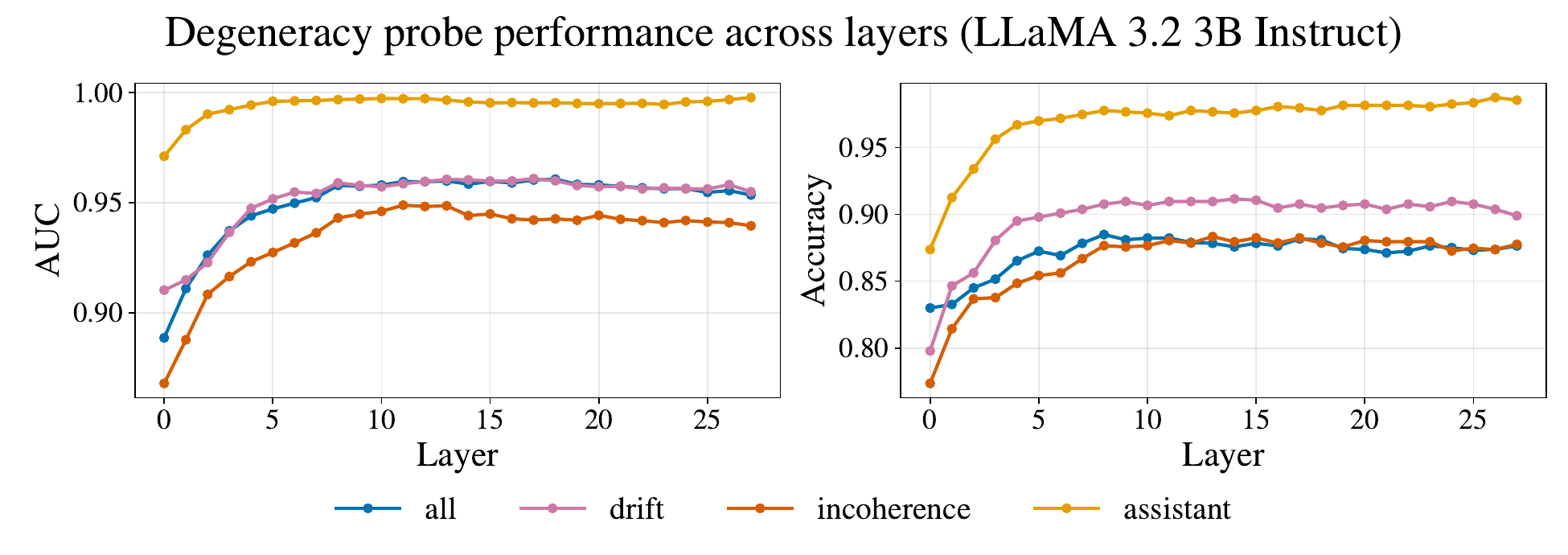}
  \caption{Degeneration probe performance at every layer of Llama-3.2-3B-Instruct: AUC (left) and accuracy (right) for the overall probe and each class probe.}
  \label{fig:degenprobelayers}
\end{figure}

Degeneracy probes are logistic regressions on mean-pooled hidden states, trained per layer with 5-fold cross-validation grouped by run (Figure~\ref{fig:degenprobelayers}). The genuine class consists of both Conrad's 529 paragraphs and the 250 genuine LLM generations because a probe trained with only clean LLM generations as the genuine side calls 70\% of Conrad's actual prose degenerate; the corresponding AUCs against clean generations only are 0.86--0.99.

\subsection{Degeneracy Steering Vector Configuration Selection}\label{app:degenvectors}
\begin{figure}[h]
  \centering
  \includegraphics[width=\textwidth]{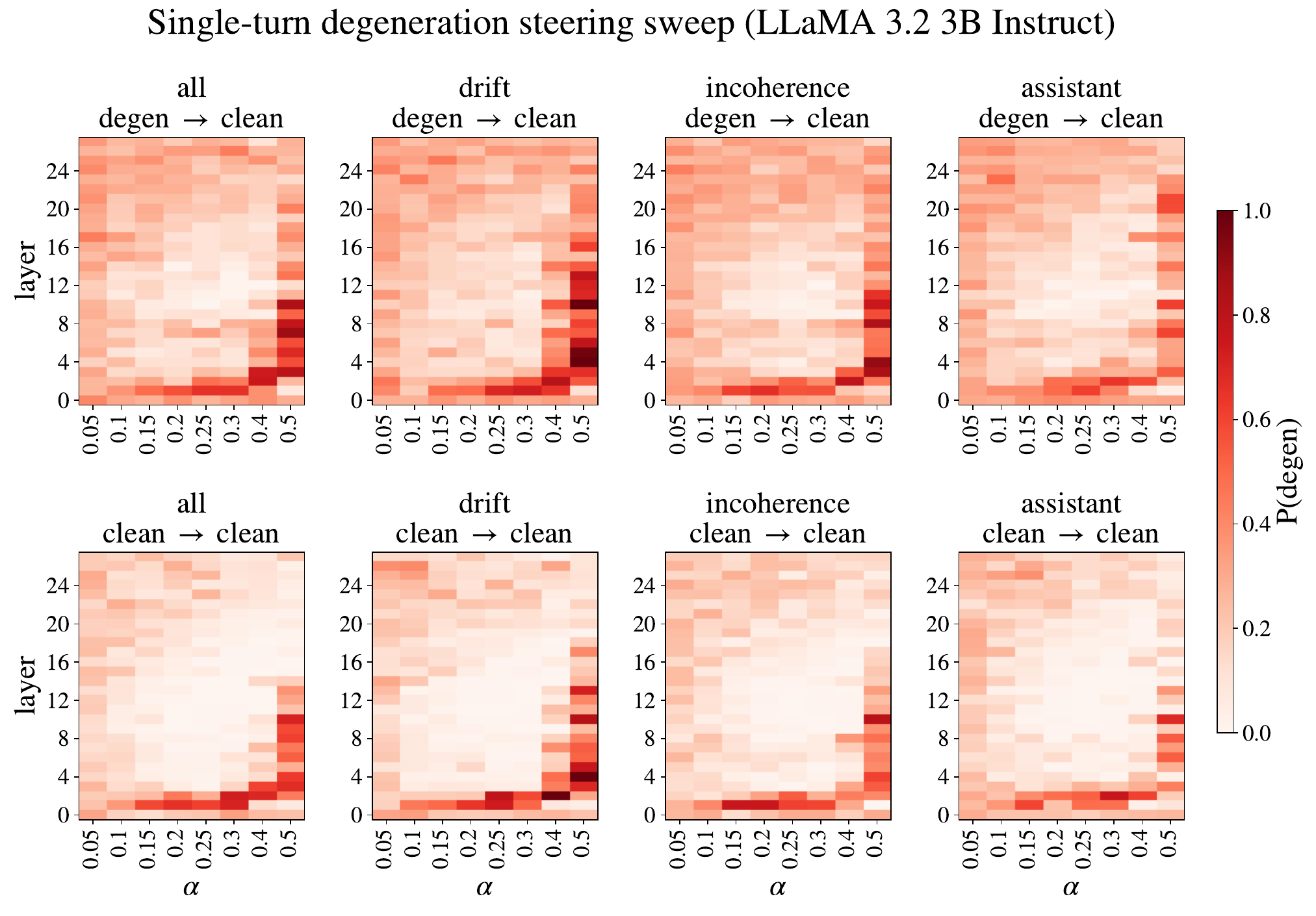}
  \caption{The degeneration steering sweep: mean $P(\mathrm{degen})$ of the continuation by layer and relative strength $\alpha$, for each vector, continuing from degenerate seeds (top row) and clean seeds (bottom row).}
  \label{fig:degensweepheat}
\end{figure}

\begin{figure}[h]
  \centering
  \includegraphics[width=\textwidth]{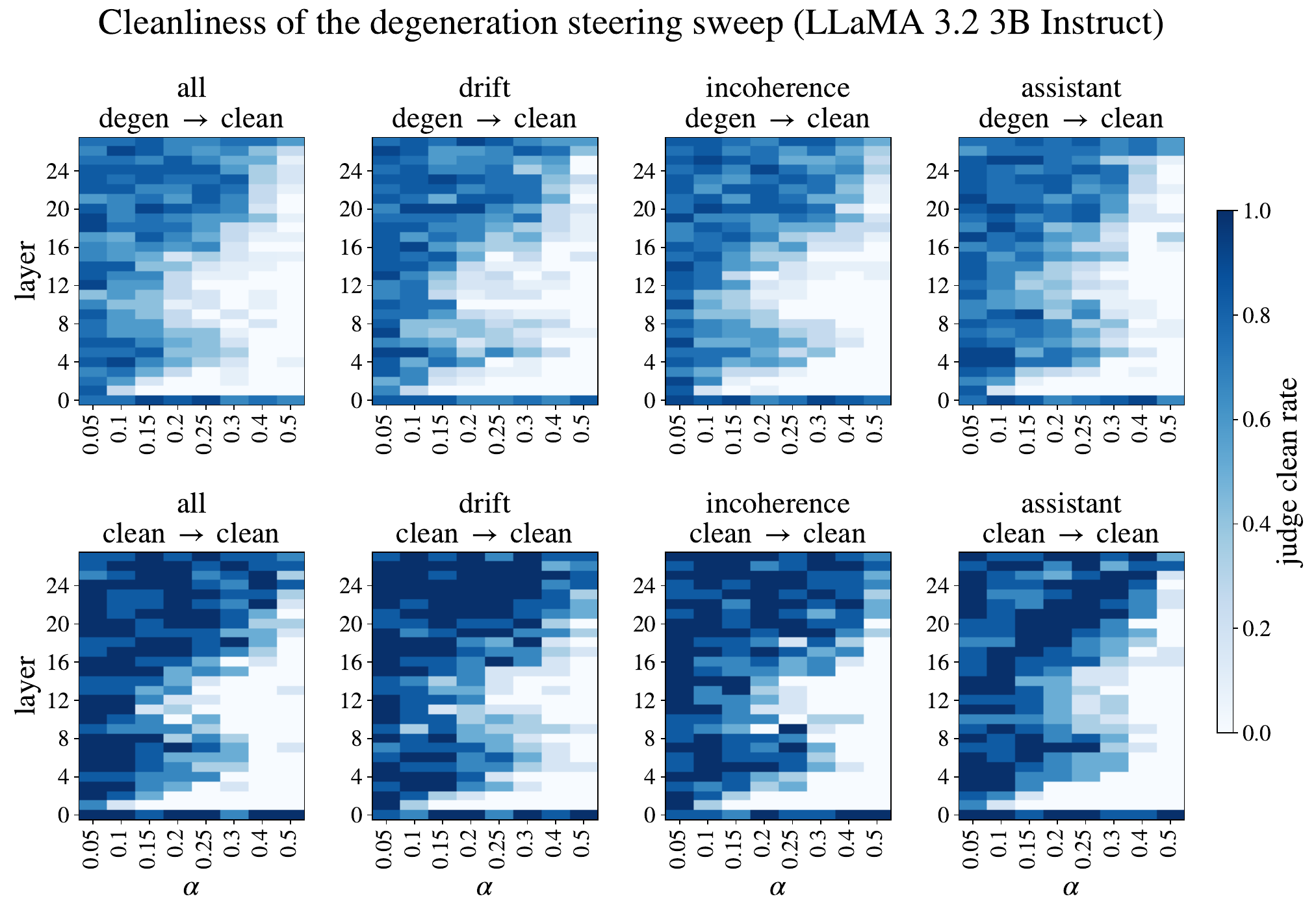}
  \caption{Judged cleanliness of the degeneration steering sweep: fraction of continuations the degeneration judge rates clean.}
  \label{fig:degenjudgeheat}
\end{figure}

Each degeneracy vector is the difference of mean last-token activations between Conrad's 529 paragraphs and a class's 250 degenerate passages; the all-degeneracy vector draws its 250 equally from the three classes. We sweep each of the four vectors (all-degeneracy, drift, incoherence, and assistant) over all 28 layers and the same 8 strengths $\alpha \in [0.05, 0.5]$, generating single-turn continuations from 18 held-out seed passages: four severity-ranked degenerate seeds per family and six clean seeds, all disjoint from the probe training data (16,128 continuations). From the probe (Figure~\ref{fig:degensweepheat}), strong mid-layer steering appears to succeed from every seed. From the judge (Figure~\ref{fig:degenjudgeheat}), cleanliness declines monotonically with $\alpha$ --- on degenerate seeds, from 78\% at $\alpha = 0.05$ to 10\% at $\alpha = 0.5$ --- against unsteered references of 83\% (degenerate seeds) and 94\% (clean seeds). We therefore select, per vector, from the configurations whose judged cleanliness stays at or near the unsteered reference on both degenerate and clean seeds, guided by probe score: all-degeneracy at layer 12, $\alpha = 0.05$; drift at layer 21, $\alpha = 0.25$; incoherence at layer 24, $\alpha = 0.20$; assistant at layer 4, $\alpha = 0.10$.

\subsection{Degeneracy Experiments}\label{app:degenexp}
\paragraph{Recovery from Degenerate Context.} For the recovery experiments, we construct the data by sampling contexts at every level of degeneracy. These contexts are consecutive, non-overlapping ten-passage windows drawn from the judged runs and kept exactly as generated. We define contamination as the number of judge-flagged passages $k$. The main set consists of 200 contexts per $k = 0, \dots, 10$, drawn round-robin across the judged $d=1$ runs. For per-class comparisons, we build one set per class from all judged runs: contexts whose flagged passages are at least two-thirds one class, up to 50 per contamination level $k \in \{2, 4, 6, 8, 10\}$, resulting in 250 drift, 245 incoherence, and 137 assistant contexts.

From each context, we generate one continuation per intervention: \textit{no intervention}; \textit{probe + resample} (regenerate while $P(\mathrm{degen}) > 0.5$, up to 8 attempts, else emit the least degenerate); \textit{probe + steer} (identical, but retries add the steering vector); and \textit{steer always} (a single steered generation, no probe in the loop), using the configurations of \S\ref{app:degenvectors}. The outcome is whether the LLM judge rates the continuation clean. 

\begin{figure}[h]
  \centering
  \includegraphics[width=\textwidth]{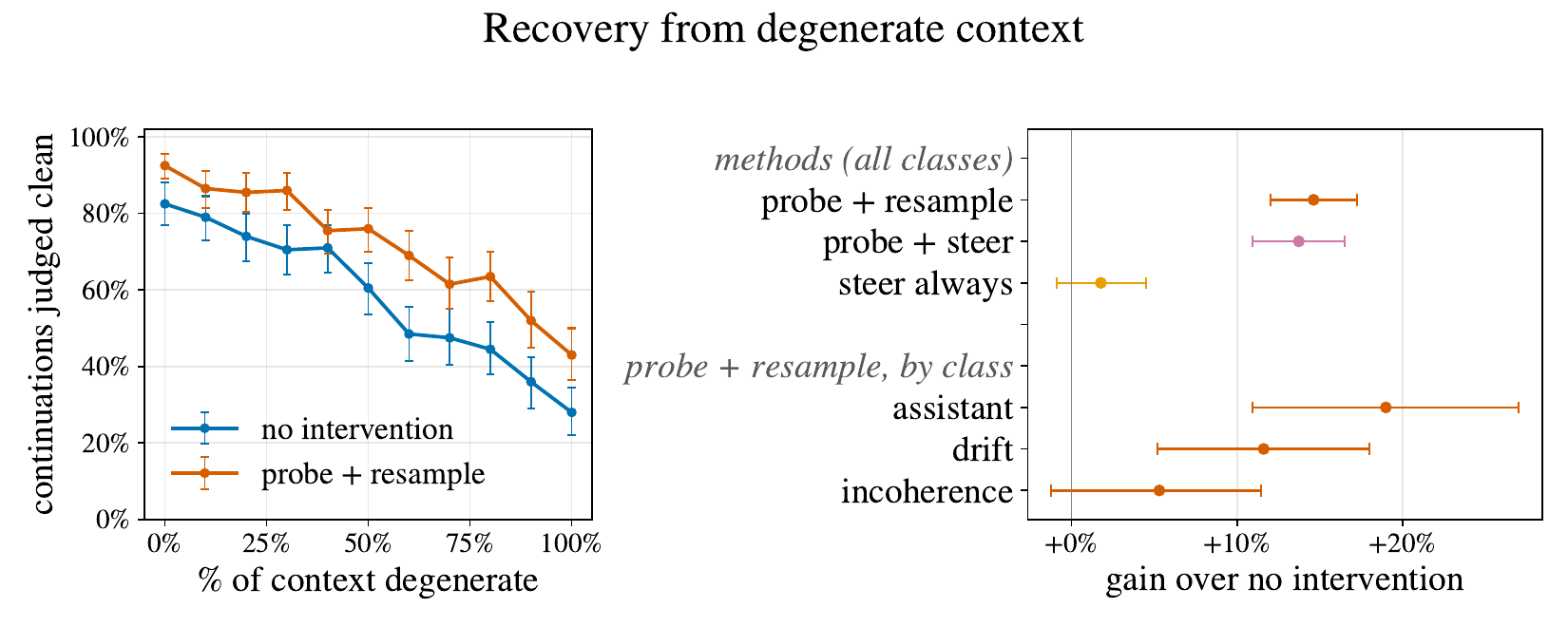}
  \caption{Recovery from degenerate context. Left: continuations judged clean vs.\ the percentage of the ten-passage context that is degenerate, comparing no intervention with probe + resample (200 contexts per level). Right: gain over no intervention, paired on the same contaminated contexts, with 95\% CIs; bottom rows split probe + resample by degeneracy class.}
  \label{fig:recovery}
\end{figure}

Figure~\ref{fig:recovery} shows results of recovery over degenerate context across the methods and per-class experiments. Paired on the same contaminated contexts ($k \ge 2$, $n = 1{,}800$), resampling gains $+14.6$ points with 95\% CI $[+12.0, +17.3]$, steered retries $+13.7$ $[+10.9, +16.4]$, the vector adds nothing over detection, and always-on steering $+1.8$ $[-0.9, +4.4]$. The gain is largest on assistant contamination ($+19.0$), in the middle on drift ($+11.6$), and smallest on incoherence ($+5.3$). The $0.5$ threshold for the degeneracy probe is validated: continuations the probe scores below it are judged clean at a roughly constant 80\%, and above it the rate falls steeply.

\begin{figure}
  \centering
  \includegraphics[width=\textwidth]{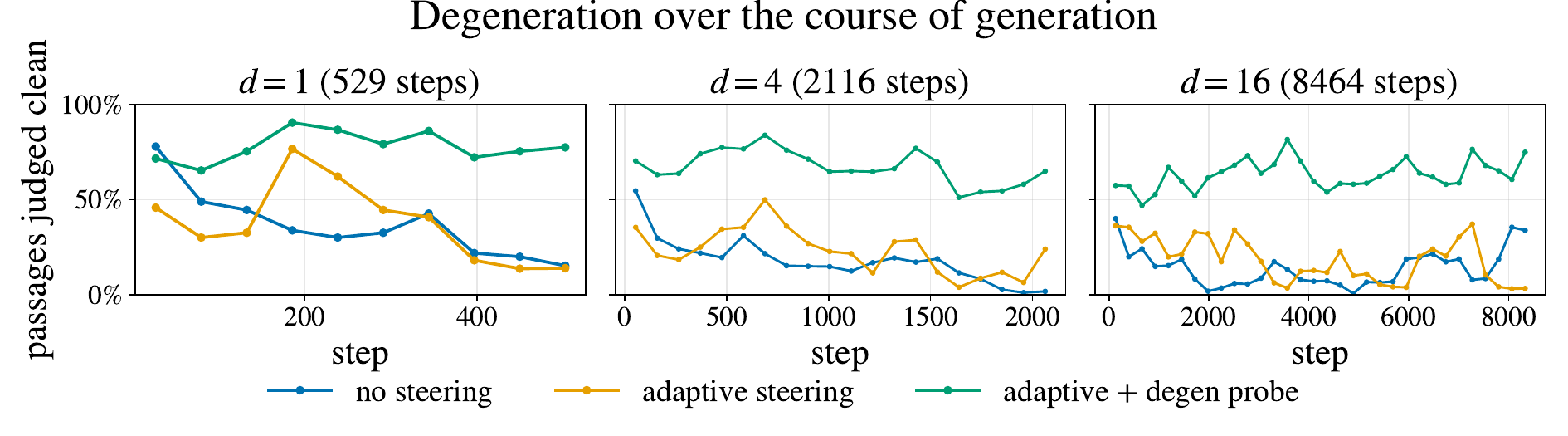}
  \caption{Degeneration over the course of generation at resolutions ($d \in \{1, 4, 16\}$): percentage of passages the judge rates clean, by step (three seeds pooled) at context window 100. The $d=16$ panel is also shown in Figure~\ref{fig:degencombined}.}
  \label{fig:degenovertime}
\end{figure}

\paragraph{Degeneration over the Course of Generation.} In Figure~\ref{fig:degenovertime}, we add the overall degeneracy probe to adaptive steering's selection loop: both probes read the same cached hidden states, candidates with $P(\mathrm{degen}) > 0.5$ are rejected by the degeneracy probe, and Beaufort selection runs on the survivors. We run this at $d \in \{1, 4, 16\}$ at context window 100. The two compressed resolutions ($d = 1/4$ and $d = 1/16$) are too short for this time course. The degeneracy probe rejection removes 3.0--3.4 of 8 candidates per step depending on resolution. When all eight are flagged (0.2--1.1\% of steps), we emit the least degenerate candidate. 

We ran a variant at $d \in \{1, 4\}$ that additionally steers these all-flagged batches with the degeneracy vector and found that it is indistinguishable from the degeneracy probe alone, consistent with the recovery result that the vector adds nothing over detection.

The Beaufort arc adherence takes a hit at calm targets. At $d=1$, the mean absolute error between probe score and the Beaufort target (on the probe's $[0,1]$ scale) is 0.19--0.32 at forces $\le 2$, against 0.03--0.15 from pure adaptive steering, while at forces $\ge 4$ the two conditions differ by at most 0.08. Drifted prose reads as calm to the Beaufort probe, so rejecting drift candidates thins exactly the calm end of the candidate pool. At $d = 16$, the runs with the degeneracy probe have 19--50\% of passages judged degenerate, against 79--92\% without it (across no steering and adaptive steering). At every resolution, the two ranges do not overlap: the worst run with the degeneracy probe is cleaner than the best run without it.

\section{Surfaced Biases}\label{app:biases}
Beyond the degeneration taxonomy, the generations surface demographic and social biases. Passages mentioning women (wives, daughters, governesses) emerge at the calm end of the arc. This is present even in unsteered runs, and steering toward calm visibly amplifies it. Additionally, Chinese people appear in stereotyped roles: the novella's two hundred Chinese laborers, whom Conrad's period narration already renders as near-cargo, are recast by the model with its own tropes --- docile contentment, superstition, and, when the model invents named Chinese characters (Wong, Chen, Lin), they often cook, serve tea, and keep the galley while the named officers command, though invented Chinese officers and engineers appear as well. Table~\ref{tab:biases} provides examples of these biases.

\begin{table}[h]
  \centering
  \caption{Examples of demographic and social biases surfacing in the generations. In the second example the model invents a wife named Alice.}
  \label{tab:biases}
  \footnotesize
  \setlength{\tabcolsep}{4pt}
  \begin{tabular}{@{}>{\raggedright\arraybackslash}p{0.16\textwidth}p{0.73\textwidth}@{}}
    \toprule
    Bias & Example \\
    \midrule
    Women appear with calm &
    \textit{As Captain MacWhirr continued to converse with Madame Boulger, his expression remained conservative, his eyes, however, betrayed a hint of interest in her account of his daughter's education.\dots he assessed her suitability for the position he was considering.} \\
    \addlinespace
    &
    \textit{As he reviewed the plans of his upcoming trip to a southern province, Captain MacWhirr's thoughts drifted to his wife, Alice, who had often expressed concern about his tendency to devote time to intellectual pursuits rather than devoting sufficient attention to family matters.} \\
    \addlinespace
    Chinese people in stereotyped roles &
    \textit{\dots nodding to the familiar faces of the Chinese coolies, who worked diligently at their assigned tasks, keeping a watchful eye on their belongings, which consisted mostly of simple clothing and a few trinkets. They smiled quietly to themselves as they went about their duties\dots} \\
    \addlinespace
    &
    \textit{\dots a sense of tension had begun to build among the Chinese coolies, fueled by superstitions about the impending typhoon season. Many believed that the gods disapproved of their treatment and livelihood\dots} \\
    \addlinespace
    &
    \textit{In the fading twilight, the flickering flames danced across Chen's weathered features\dots nostalgia flooding his eyes as Lin ladled hot stew into worn tin mugs.} \\
    \bottomrule
  \end{tabular}
\end{table}

\section{Audio-Visual Installation}
\subsection{Voice Cloning}\label{app:voiceclone}
To source audio samples for voice cloning, the artist recorded 8 speech samples (4 calm, 4 storm), each increasing in intensity, reading from the same script. We flag degenerate passages with the degeneracy probe. Flagged degenerate passages are cloned with Qwen3-TTS from the recording whose 0.125-wide intensity bucket contains the passage's Beaufort probe score. Qwen3-TTS proved more robust than Chatterbox-Turbo when synthesizing repetitive and degenerate inputs, which is why such passages are routed to it.

\subsection{Audio Post-Processing}\label{app:audiopp}
All synthesized audio is post-processed with voice activity detection \cite{SileroVAD}, a light effects chain (equalization, compression, reverb, limiting), and loudness normalization \cite{steinmetz2021pyloudnorm} to reduce potential TTS artifacts (e.g. extended silences) and ensure consistent loudness across the whole experience.

\begin{table}[t]
  \centering
  \caption{Real-time factor (lower is better) of each stage of the installation pipeline on an RTX A5000, with each stage's share of the total.}
  \label{tab:rtf}
  \small
  \begin{tabular}{lcc}
    \toprule
    Stage & RTF & Share of pipeline \\
    \midrule
    LLM batched steering        & 0.145 [0.144--0.147] & 29.5\% \\
    Probe scoring               & 0.004 [0.003--0.004] & 0.7\% \\
    TTS: Chatterbox-Turbo       & 0.329 [0.326--0.330] & 66.5\% \\
    Audio post-processing       & 0.016 [0.016--0.016] & 3.3\% \\
    \midrule
    Full pipeline               & 0.494 [0.490--0.497] & 100\% \\
    \bottomrule
  \end{tabular}
\end{table}

\subsection{Real-Time Performance}\label{app:rtf}
We measure the real-time factor (RTF) of each stage of the installation (Table~\ref{tab:rtf}), under the installation's generation configuration (\S\ref{implementation}: context window $w=10$, 150 new tokens per passage). We report results on the RTX A5000 (24\,GB). Each run generates 60 sections along the full Beaufort arc; we discard the first 30 to let the context window saturate, since the cost of steering grows with context. We report the median and [min--max] over 6 runs with distinct sampling seeds, spread across 3 machines. The whole pipeline runs at 0.49, comfortably faster than real-time. Degenerate passages routed to Qwen3-TTS synthesize at 0.341 on the same hardware, in place of the Chatterbox-Turbo stage, still resulting in a faster than real-time system.


\newpage

\end{document}